\documentclass[journal]{IEEEtran}
\usepackage{subfig}
\usepackage{graphicx}
\usepackage{amsmath}
\usepackage{amssymb}
\usepackage{booktabs}
\usepackage{bbding}
\usepackage{multirow}
\usepackage{float}
\usepackage{xcolor}
\usepackage{colortbl,booktabs}
\usepackage{color}
\usepackage[pagebackref,breaklinks,colorlinks]{hyperref}
\definecolor{tablegreen}{RGB}{113,198,113}
\definecolor{tablered}{RGB}{205,51,51}
\definecolor{linkcolor}{HTML}{ED1C24}
\definecolor{baselinecolor}{gray}{.9}
\newcommand{\baseline}[1]{\cellcolor{baselinecolor}{#1}}

\ifCLASSINFOpdf
\else
\fi
\begin{document}
%
\title{StreamYOLO: Real-time Object Detection for Streaming Perception}
%
%
%

\author{Jinrong~Yang,
        Songtao~Liu,
        Zeming~Li,
        Xiaoping~Li,
        and~Jian~Sun
\thanks{J. Yang and X. Li are with the State Key Laboratory of Digital Manufacturing Equipment and Technology, Huazhong University of Science and Technology, Wuhan 430074, China e-mail: (see yangjinrong@hust.edu.cn; lixiaoping@hust.edu.cn)}
\thanks{S. Liu, Z. Li and J. Sun are with the Megvii Technology, Beijing, 100080, China e-mail: (see liusongtao@megvii.com; lizeming@megvii.com; sunjian@megvii.com).}
\thanks{Manuscript received xxx; revised xxx.}}

%
%

\markboth{Journal of \LaTeX\ Class Files,~Vol.~14, No.~8, August~2015}%
{Shell \MakeLowercase{\textit{et al.}}: Bare Demo of IEEEtran.cls for IEEE Journals}
%



\maketitle

\begin{abstract}
The perceptive models of autonomous driving require fast inference within a low latency for safety. While existing works ignore the inevitable environmental changes after processing, streaming perception jointly evaluates the latency and accuracy into a single metric for video online perception, guiding the previous works to search trade-offs between accuracy and speed. In this paper, we explore the performance of real time models on this metric and endow the models with the capacity of predicting the future, significantly improving the results for streaming perception. Specifically, we build a simple framework with two effective modules. One is a Dual Flow Perception module (DFP). It consists of dynamic flow and static flow in parallel to capture moving tendency and basic detection feature, respectively. Trend Aware Loss (TAL) is the other module which adaptively generates loss weight for each object with its moving speed. Realistically, we consider multiple velocities driving scene and further propose Velocity-awared streaming AP (VsAP) to jointly evaluate the accuracy. In this realistic setting, we design a efficient mix-velocity training strategy to guide detector perceive any velocities. Our simple method achieves the state-of-the-art performance on Argoverse-HD dataset and improves the sAP and VsAP by 4.7\% and 8.2\% respectively compared to the strong baseline, validating its effectiveness. 
\end{abstract}

\begin{IEEEkeywords}
Streaming perception, object detection, video prediction, mix-velocity training.
\end{IEEEkeywords}

%
\IEEEpeerreviewmaketitle

\section{Introduction}
%
%
%
%

\IEEEPARstart{D}{ecision-making} biases and errors that occur in autonomous driving will expose human lives be at stake. Wrong decisions often become more serious as the velocity of the vehicle increases. To avoid the irreversible situation, one effort is to perceive its environment and (re)act within a low latency. 

\begin{figure}[t]
\centering      
\subfloat[baseline]
{\includegraphics[width =1.7in]{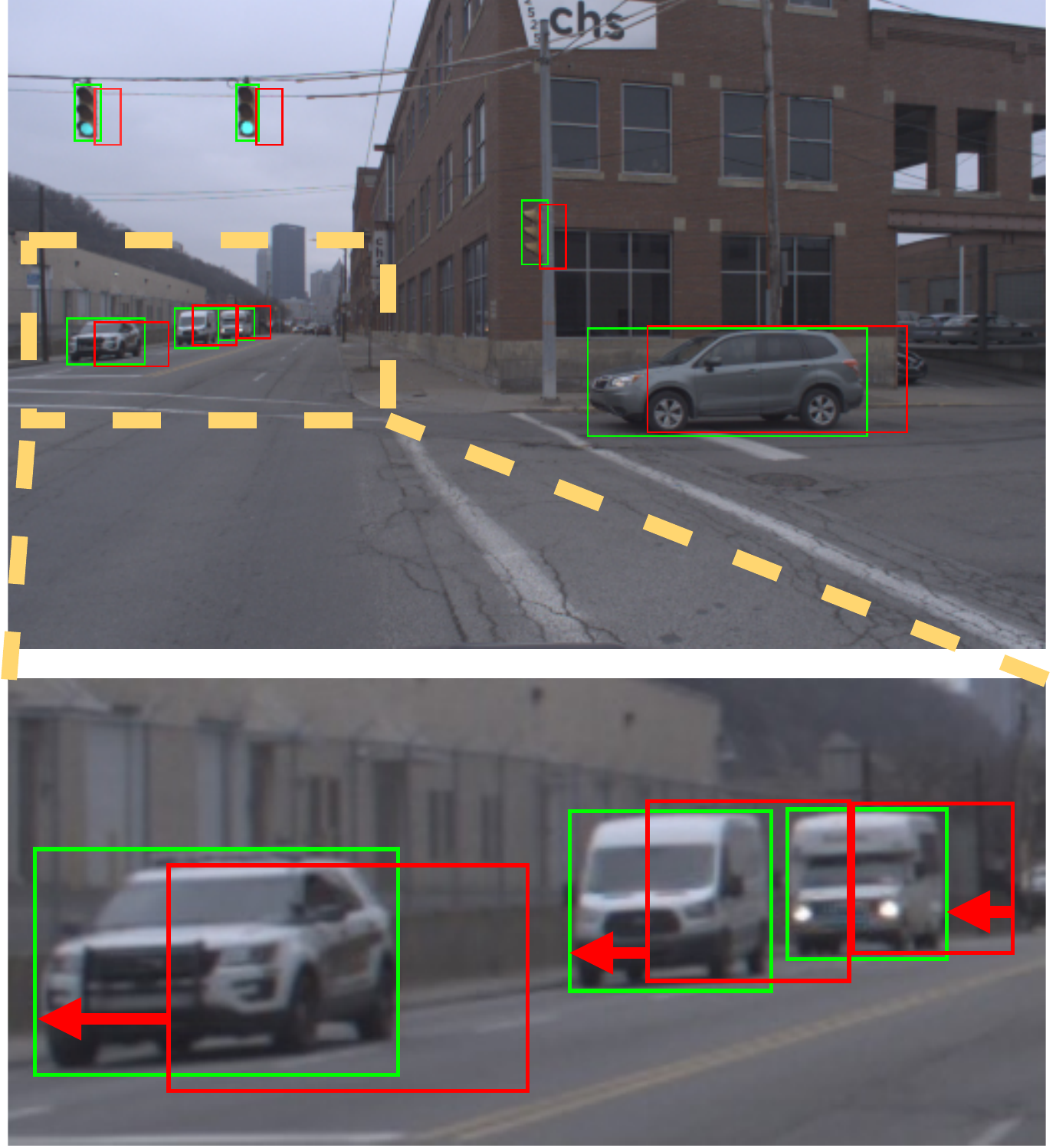}
\label{fig:base}}
\subfloat[ours]
{\includegraphics[width = 1.7in]{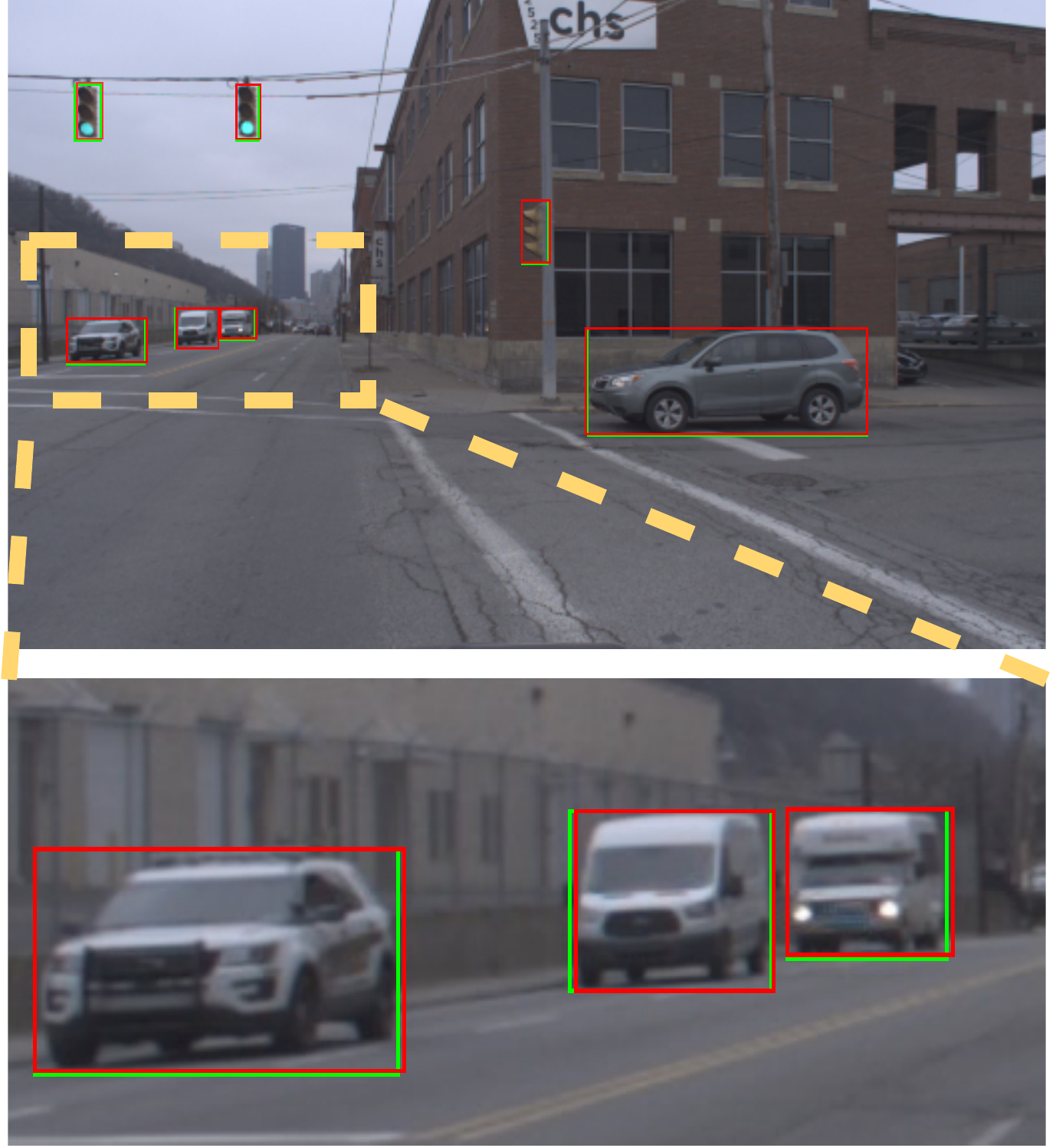}
\label{fig:ours}}\\      
\subfloat[Different velocities]
{\includegraphics[width = 3.45in]{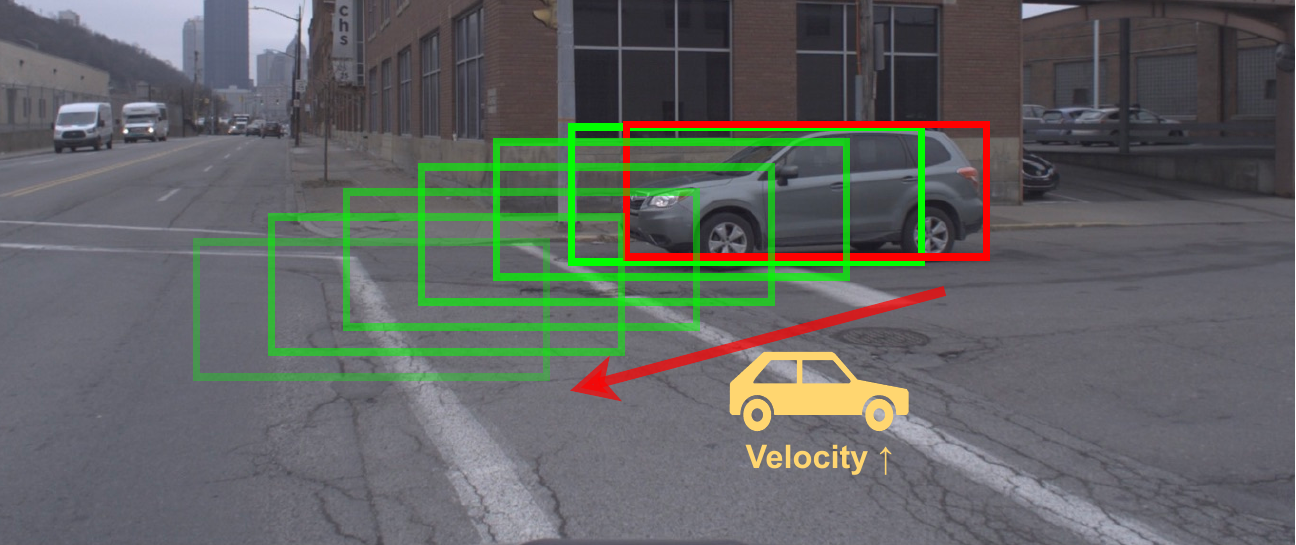}
\label{fig:velocity}}\\ 

\caption{Illustration of visualization results of base detector and our method. The green boxes are ground truth, while the red ones are predictions. The red arrows mark the shifts of the prediction boxes caused by the processing time delay while our approach alleviates this issue. The lighter the color of the red box, the greater the offset caused by the faster speed.}      
\label{fig:fig1}      
\vspace{-0.1in}  
\end{figure}

Of late, several real-time detectors ~\cite{yolo1,yolo2,yolo3,yolo4,yolo5,yolox} with strong performance and low latency constraints have come to the fore. However, they are yet mainly studied in an \emph{offline} setting ~\cite{streamer}. In a real-world online scenario, no matter how fast the algorithm gets, once the algorithm has handled the latest observations, the state of the world around the vehicle will change. As shown in Fig.~\ref{fig:base}, deviations between the altered state and the perceived results may trigger unsafe decisions for autonomous driving. As the velocity increased, the problem was further exacerbated (see Fig.~\ref{fig:velocity}). Therefore, for both online perception and the real world, algorithms are expected to perceive future world states in relation to actual processing time.


To address this issue, ~\cite{streamer} first proposes a new metric dubbed streaming accuracy, which coherently integrates accuracy and latency into a single metric for real-time online perception. It jointly evaluates the results of the entire perception stack at each moment, forcing the algorithm to perceive the state of the model to complete the processing.

Under this practical evaluation framework, several strong detectors ~\cite{htc,retinanet,maskrcnn} show significant performance degradation~\cite{streamer} from offline setting to streaming perception. Going one step further, ~\cite{streamer} proposes a meta-detector named Streamer that works with decision-theoretic scheduling, asynchronous tracking, and Kalman filter-based prediction to rescue much of the performance degradation. Following this work, Adaptive streamer~\cite{adaptivestreamer} employs various approximate executions based on deep reinforcement learning to learn a better trade-off online. These works mainly seek a better trade-off plan between speed and accuracy in existing detectors, while the design of a novel streaming perception model has not been well explored. Unlike these, Fovea~\cite{fovea} focuses on the ability to utilize KDE-based mapping techniques to help detect small objects. The effort actually improves the performance of the meta-detector, but fails to address the intrinsic problem of streaming perception.

One thing that the above works ignore is the existing real-time object detectors ~\cite{yolo5, yolox}. With the addition of powerful data augmentation and a refined architecture, they achieve competitive performance and can run faster than 30 FPS. With these "fast enough" detectors, there is no room to trade off accuracy versus latency on streaming perception, since the detector's current frame result is always matched and evaluated by the next frame. These real-time detectors can narrow the performance gap between streaming perception and offline settings. Indeed, both the 1st ~\cite{yoloxx} and 2nd ~\cite{2nd} place solutions of Streaming Perception Challenge (Workshop on Autonomous Driving at CVPR 2021) utilise real-time models YOLOX ~\cite{yolox} and YOLOv5 ~\cite{yolo5} as their base detectors.

Standing on the shoulders of the real-time models, we find that the performance gaps now all come from the fixed inconsistency between the current processing frame and the next matching frame. Therefore, the key solution for streaming perception is switched to forecast the results of the \emph{next} frame in the \emph{current} state. 

Through extensive experiments, we also found that the upper limit of performance drop is absolutely dependent on the driving velocity of the vehicle. As show on Fig.~\ref{fig:velocity}, the quicker the velocity is, the larger the performance drop. As a special case, no performance drop when stationary. This phenomenon is reasonable as driving at high speeds is prone to traffic accidents. Therefore, velocity perception is an essential part of forecasting \emph{consistency} results.
 
Unlike the heuristic methods such as Kalman filter ~\cite{kalman} adopted in ~\cite{streamer}, in this paper, we directly endow the real-time detector with the ability to predict the future of the next frame. Specifically, we construct triplets of the last, the current, and next frame for training, where the model takes the last and current frames as input and learns to predict the detection results of the next frame. 

To improve the training efficiency, we propose two key design schemes: i) For model architecture, we conduct a Dual-Flow Perception (DFP) module to fuse the feature map from the last and the current frames. It consists of a dynamic flow and a static flow. Dynamic flow focuses on the motion trend of objects to predict while static flow provides basic information and features of detection through a residual connection. ii) For the training strategy, we introduce a Trend Aware Loss (TAL) to dynamically assign different weights to locate and predict each object, since we find that objects within one frame may have different moving speeds. 

In the realistic driving scene, vehicles often drive at different velocities including static. Therefore, training detector to perceive the future at a relatively fixed velocity range may cause to be unable to make the correct decisions at different velocities. To this end, we design a mix-velocity training strategy to efficiently train our detector to better perceive any velocities. Finally, we further design Velocity-aware streaming AP (VsAP) metric to coherently evaluate sAP on all velocity settings.

We conduct comprehensive experiments on Argoverse-HD~\cite{argoverse,streamer} dataset, showing significant improvements in the stream perception task. 
In summary, the contributions of this work are as four-fold as follows:
\begin{itemize}
\item With the strong performance of real-time detectors, we find the key solution for streaming perception is to predict the results of the \emph{next} frame. This simplified task is easy to be structured and learned by a model-based algorithm.  

\item We build a simple and effective streaming detector that learns to forecast the next frame. We propose two adaptation modules, \emph{i.e.}, Dual-Flow Perception (DFP) and Trend Aware Loss (TAL), to perceive the moving trend and predict the future.      

\item We find perceiving the driving velocity is the key to forecasting the consistently future results, and propose a mix-velocity training approach to endow model the ability to tackle different driving velocities. We also propose VsAP to evaluate the performance in all velocity setting.

\item We achieve competitive performance on Argoverse-HD~\cite{argoverse,streamer} dataset without bells and whistles. Our method improves the sAP by +4.7\% (the VsAP by 8.2\%) compared to the strong baseline of the real-time detector and shows robust forecasting under the different moving speeds of the driving vehicle. 
\end{itemize}

Some preliminary results appear in ~\cite{yang2022real}, and this paper includes them but significantly extends in the following aspects.

\begin{itemize}
\item We point out that training our detector with only using a single velocity scale will fail to perceive the future at differently driving speeds. To this end, we propose a mix-velocity training strategy to enhance the model to tackle different driving velocities.

\item We find that random horizontal flip is the key for the detector to perceive the clue of the relative driving direction. Therefore, we update all experiments to establish a more strong detector.

\item We conduct more experiments to establish our strong pipeline (ie., sample assignment of anchor points, more fusion operators, more prediction tasks, pre-training policies, and data augmentations) and ablate more factors to verify the superiority of our method (vehicle category, quicker driving speed, image scale, etc.).

\end{itemize}

\section{Related Works}
\label{sec:related}

\textbf{Image object detection.} In the era of deep learning, detection algorithms can be divided into the two-stage ~\cite{fastrcnn,fpn,fasterrcnn} and the one-stage ~\cite{ssd,retinanet,fcos,yolo1} frameworks. The two-stage networks jointly train RPN and fast R-CNN to perform a paradigm that first finds objects and then refines classification and regression. The coarse-to-fine manner shows significant performance but undergoes larger latency. One-stage detectors abandon the proposal generation stage and directly output bounding boxes with final positioning coordinates and class probabilities. The one-step paradigms endow one-stage models with higher inference speed than two-stage ones. With the increasingly stringent requirements for model inference speed in reality, many works such as the YOLO series ~\cite{yolo1,yolo2,yolo3,yolo4,yolo5,yolox} focus on pursuing high performance and optimizing inference speed. These high speed models are equipped with a bunch of advanced techniques, including efficient backbones~\cite{cspnet,pan}, strong data augmentation strategies~\cite{mixup,copypaste}, etc. Our work is based on the state-of-the-art real-time detector YOLOX ~\cite{yolox} which achieves strong performance among real-time detectors. 

\textbf{Video object detection.} Streaming perception also involves video object detection (VOD). The VOD task aims to alleviate the poor situation caused by the complex video variations, \emph{e.g.}, motion blur, occlusion, and out-of-focus. Attention-based methods~\cite{mega,rdn} employ attention mechanism~\cite{rnl} to establish a strong information (context, semantics, etc.) association between the local temporal information of key and support frames. Flow-based methods utilize optical flow to propagate features from key frames to support frames~\cite{dff} or to extract the temporal-spatial clues between key frames and support ones~\cite{fgfa}. LSTM-based methods~\cite{lstm1,lstm2,lstm3} use convolutional long short term memory (LSTM~\cite{lstm}) to process and save important temporal-spatial clues and enhance key frames. Tracking-based methods~\cite{tracker, tracker2} make use of the motion prediction information to detect interval frames adaptively and track frames in between. These frameworks focus on improving the detection accuracy for key frames, which needs to use history or future frames to conduct an enhanced process. They all focus on the offline setting while streaming perception considers the online processing latency and needs to predict the future results.

\textbf{Streaming perception.} Streaming perception task coherently considers latency and accuracy. ~\cite{streamer} firstly proposes streaming AP (sAP) to evaluate accuracy under the consideration of time delay. Facing latency, non-real-time detectors will miss some frames. ~\cite{streamer} proposes a meta-detector to alleviate this problem by employing Kalman filter~\cite{kalman}, decision-theoretic scheduling, and asynchronous tracking~\cite{tracker}. ~\cite{adaptivestreamer} lists several factors (e.g., input scales, switchability of detectors, and scene aggregation.) and designs a reinforcement learning-based agent to learn a better combination for a better trade-off. Fovea~\cite{fovea} employ a KDE-based mapping to magnify the regions of small objects (e.g., long-distance vehicles appear in the image) so that the detectors can detect small objects without feeding larger resolution image. In short, ~\cite{streamer} and ~\cite{adaptivestreamer} focus on searching for a better trade-off by utilizing Non-learning and learnable strategies, while Fovea~\cite{fovea} actually improves the upper performance of the detector. Instead of utilizing non-real-time detectors, both the 1st~\cite{yoloxx} and 2nd~\cite{2nd} place solution of Streaming Perception Challenge (Workshop on Autonomous Driving at CVPR 2021) adopt real-time models (YOLOX~\cite{yolox} and YOLOv5~\cite{yolo5}) as their base detectors. By using fast enough models, the miss frames will be reduced. However, no matter how many how fast the model is, there will always be a time delay of one unit. Based on that, we establish a strong baseline with real-time detector, and simplify the steaming perception to the task of ``predicting the next frame''. And we further improve the performance by employing a learning-based model to explicitly predict the future instead of searching for a better trade-off between accuracy and latency. In addition, we explore and alleviate the streaming perception problem in case of vehicles driving at different velocities.

\textbf{Future prediction.} Future prediction tasks aim to predict the results for the unobserved future data. Current tasks are explored in the era of semantic/instance segmentation. For semantic segmentation, early works ~\cite{predicting,bayesian} construct a mapping from past segmentation to future segmentation. Recent works ~\cite{segmenting,vsaric2019single,warp,predictive} convert to predict intermediate segmentation features by employing deformable convolutions, teacher-student learning, flow-based forecasting, LSTM-based approaches, etc. For instance segmentation prediction, some approaches predict the pyramid features~\cite{luc2018predicting} or the pyramid feature of varied pyramid
levels jointly~\cite{sun2019predicting,apanet}. The above prediction methods do not consider the misalignment of prediction and environment change caused by processing latency, leaving a gap to real-world application. In this paper, we focus on the more practical task of streaming perception. To the best of our knowledge, our work is also the first method to apply future prediction to object detection, which outputs discrete set predictions compared with the existing segmentation prediction tasks.

\begin{figure}[t]
\begin{center}
\includegraphics[width=\linewidth]{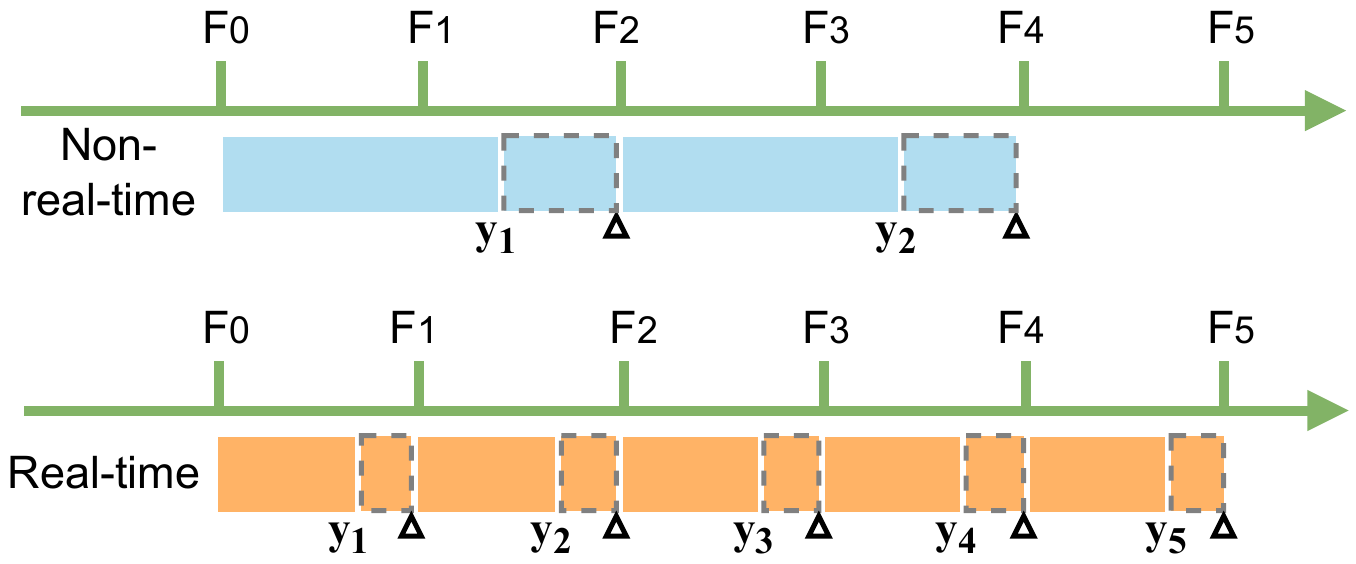}
\caption{Comparison on different detectors in streaming perception evaluation framework. Each block represents the process of the detector for one frame and its length indicates the running time. The dashed block indicates the time until the next frame data is received.}
\label{fig:streaming}
\end{center}
\end{figure}

\begin{figure}[b]
\begin{center}
\includegraphics[width=\linewidth]{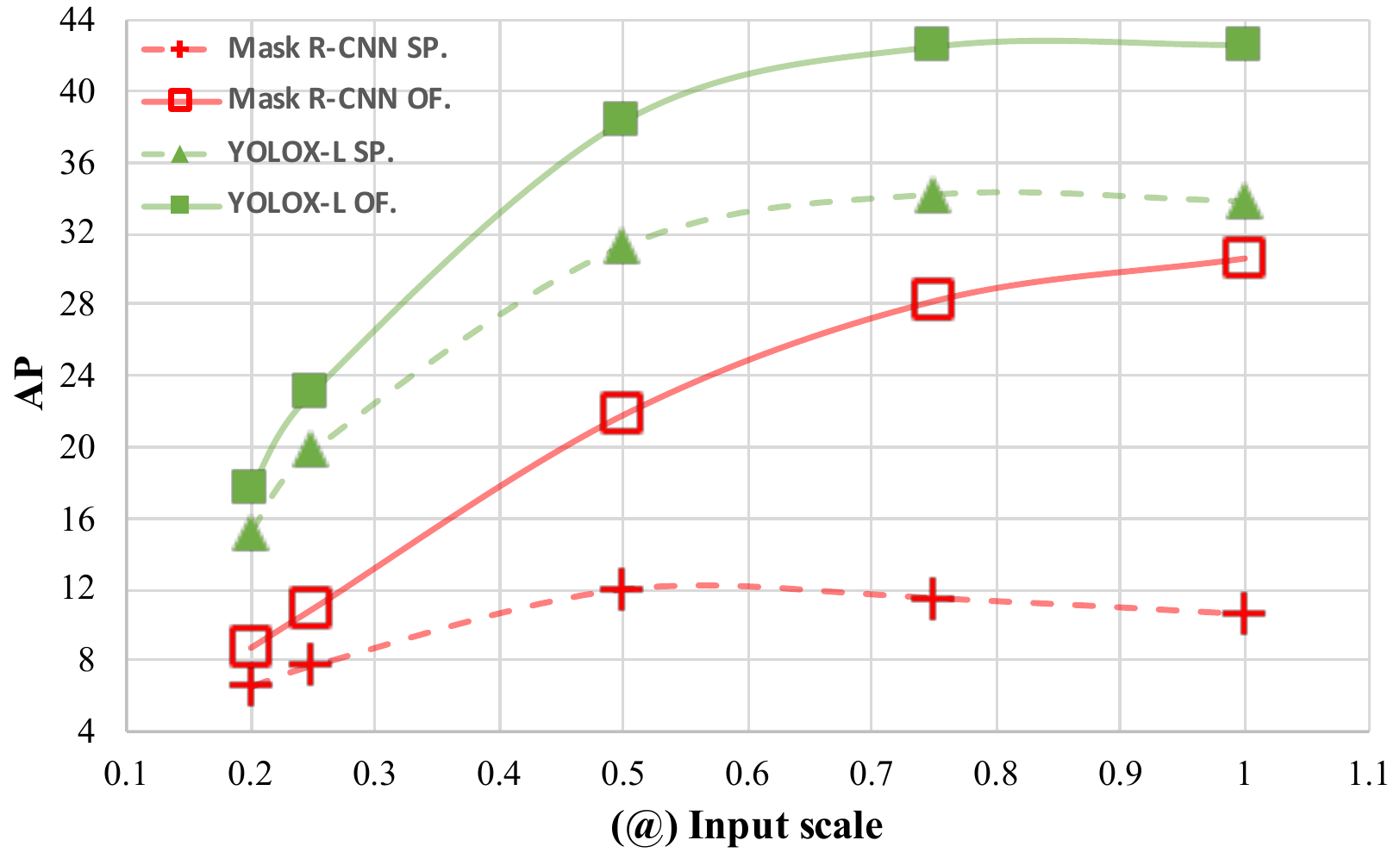}
\caption{The performance gap between \emph{offline} and streaming perception setting brings about on Argoverse-HD dataset. 'OF' and 'SP' indicate \emph{offline} and streaming perception setting respectively. The number after @ is the input scale (the full resolution is $1200 \times 1920$).}
\label{fig:contrast}
\end{center}
\end{figure}

\section{Methods}
\begin{figure*}
\setlength{\abovecaptionskip}{-3pt}
\begin{center}
\includegraphics[width=\linewidth]{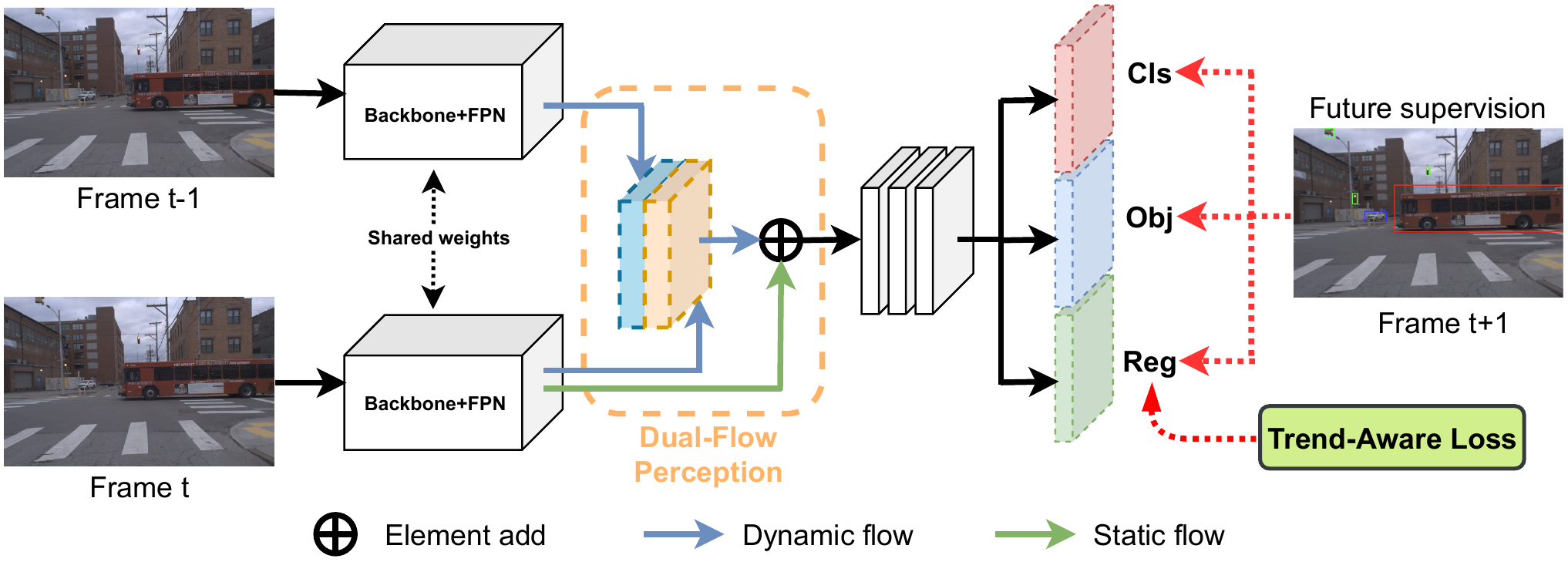}
\end{center}
\caption{The training pipeline. First, we adopt a shared weight CSPDarknet-53 with PANet to extract FPN features of the current and last image frames. Second, we use the proposed Dual-Flow Perception module (DFP) to aggregate feature maps and feed them to classification, objectness and regression head. Third, we directly utilize the ground truth of the next frame to conduct supervision. We also design a Trend-Aware Loss (TAL) applied to the regression head for efficient training.}
\label{fig:train}
\end{figure*}

\subsection{Streaming Perception}
Streaming perception organizes data as a set of sensor observations. To take the model processing latency into account, ~\cite{streamer} proposes a new metric named streaming AP (sAP) to simultaneously evaluate time latency and detection accuracy. As shown in Fig.~\ref{fig:streaming}, the streaming benchmark evaluates the detection results in a continuous time frame. After receiving and processing an image frame, sAP simulates the time latency between the streaming flow and examines the processed output with a ground truth of the real world state.

Taking a non-real-time detector as an example, the output $y_1$ of the frame $F_1$ is matched and evaluated with the ground truth of $F_3$, while the result of $F_2$ is ignored. Therefore, for streaming perception task, non-real-time detectors may miss many image frames, resulting in long-term offset results, which seriously affects the performance of \emph{offline} detection.

For real-time detectors (the total processing time of one frame is less than the time interval of image streaming), the task of streaming perception becomes explicit and easy. As shown in Fig.~\ref{fig:streaming}, the shorter the time interval the better, and if the detector takes more time than the unit time interval, a fixed pattern may not be optimal as shown in the decision-theoretic analysis in ~\cite{streamer}, therefore, a fast real-time detector avoids these problems. This fixed matching pattern not only eradicates the missed frames but also reduces the time shift for each matched ground truth. 

In Fig.~\ref{fig:contrast}, we compare two detectors, Mask R-CNN~\cite{maskrcnn} and YOLOX~\cite{yolox}, using several image scales, and investigate the performance gap between streaming perception and offline settings. In the case of low-resolution inputs, the performance gap between the two detectors is small, since they both operate in real-time. However, as the resolution increases, Mask R-CNN's performance drops more because it runs slower. For YOLOX, its inference speed maintains real-time with the resolution increasing, so that the performance gap does not widen accordingly.

\subsection{Pipeline}
\label{sec:3.2}

The fixed matching pattern from real-time detectors also enables us to train a learnable model to mine latent moving trends and predict the objects of the next image frames. Our approach consists of a basic real-time detector, an offline training schedule, and an online inference strategy, which are described next.    

\paragraph{Base detector} We choose the recent proposed YOLOX ~\cite{yolox} as our base detector. It inherits and carries forward YOLO series  ~\cite{yolo1,yolo2,yolo3} to an anchor-free framework with several tricks, \emph{e.g.}, decoupled heads ~\cite{head1,head2}, strong data augmentations ~\cite{mixup,copypaste}, and advanced label assigning ~\cite{ota},  achieving strong performance among real-time detectors.
It is also the 1st place solution ~\cite{yoloxx} of Streaming Perception Challenge in the Workshop on Autonomous Driving at CVPR 2021. Unlike ~\cite{yoloxx}, we remove some engineering acceleration tricks like TensorRT and change the input scale to the half resolution ($600 \times 960$) to ensure the real-time speed without TensorRT. We also discard the extra datasets used in ~\cite{yoloxx}, \emph{i.e.}, BDD100K~\cite{yu2020bdd100k}, Cityscapes~\cite{cordts2016cityscapes}, and nuScenes~\cite{caesar2020nuscenes} for pre-training. These shrinking changes definitely degrade the detection performance compared to ~\cite{yoloxx}, but they lighten the executive burden and allow extensive experiments. We believe the shrinking changes are orthogonal to our work and can further improve performance.   

\paragraph{Future aware label assignment} We follow YOLOX to employ SimOTA~\cite{ota} for label assignment. The advanced sample policy firstly delimits a region based on object center and ground truth priors. Next, jointly combining classification and localization loss as a metric to rank all proposals. Finally, it uses a dynamic top-k strategy to choose positive samples dynamically. The overall sampling pipeline is gorgeous to train model on still task, but a fuzzy brings about in future anticipation task, i.e., how to formulate a sampling region. To this end, we study three sampling ways: current, future, and combination regions. We found that employing region based on ground truth priors of future frame achieves best optimal, which implicitly supervises model to perceive future and reduces the difficulty of regression. The comparison results are showed in Tab.~\ref{tab:sample}.

\paragraph{Training} Our overall training pipeline are shown in Fig.~\ref{fig:train}. We bind the last, the current, and the next ground truth boxes to construct a triplet $(F_{t-1}, F_t, G_{t+1})$ for training. The prime purpose for this scheme is simple and direct: it is inevitable to perceive the moving clues for each object for forecasting the future position of objects. To this end, we feed two adjacent frames ($F_{t-1}$, $F_{t}$) into model and employ the ground truth of $F_{t+1}$ to supervise model to directly predict the detection results of the next frame. Based on the manner of inputs and supervision, we refactor the training dataset to the formulation of  $\{(F_{t-1}$, $F_{t}$, $G_{t+1})\}_{t=1}^{n_{t}}$, where $n_{t}$ is the total sample number. We abandon the first and last frame of each video clip. Reconstructing the dataset as a sweet spot, as we can continue training with a random shuffling strategy and improve efficiency with normal training on distributed GPUs.

To better capture the moving trend between two input frames, we propose a Dual-Flow Perception Module (DFP) and a Trend-Aware Loss (TAL), introduced in the next subsection, to fuse the FPN feature maps of two frames and adaptively catch the moving trend for each object.

We also investigate another indirect task that predicts in parallel the current gt boxes $G_t$ and the offsets of object transition from $G_t$ to $G_{t+1}$. However, according to some ablation experiments, described in the next Sec.~\ref{exp:pred}, we find that predicting the additional offsets always falls into a suboptimal task. We also try a weak supervision manner by predicting corresponding offsets and throwing away the prediction structure in the inference stage, which also shows a suboptimal result. One of the reasons is that the transition offset value between two adjacent frames is small and contains some numerically unstable noise. It also has some bad cases where the labels of the corresponding objects are sometimes unreachable (new objects appear or the current object disappears in the next frame).

\begin{figure}[!htb]
\begin{center}
\includegraphics[width=\linewidth]{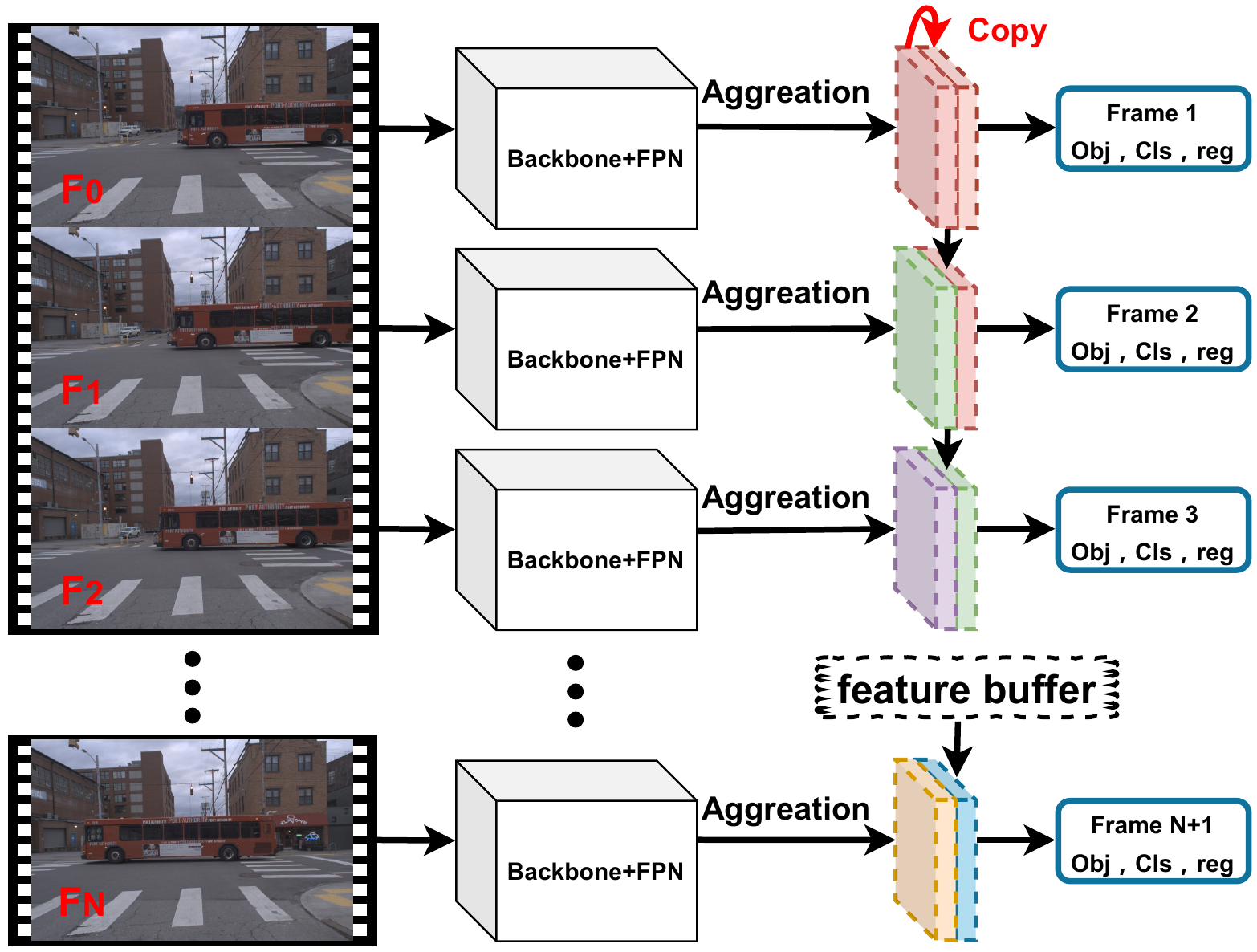}
\caption{The inference pipeline. We employ a feature buffer to save the historical features of the latest frame and thus only need to extract current features. By directly aggregating the features stored at the last moment, we save the time of handling the last frame again. For the beginning of the video, we copy the current FPN features as pseudo historical buffers to predict results.}
\label{fig:inference}
\end{center}
\end{figure}

\paragraph{Inference} The model takes two frames of images as input, which nearly doubles the computation and time consumption compared to the original detector. As shown in Fig.~\ref{fig:inference}, to remove the dilemma, we use a feature buffer to store all the FPN feature maps of the previous frame $F_{t-1}$. At inference time, our model only extracts the feature from the current image frame and then fuses it with the historical features from the buffer. With this strategy, our model runs almost as fast as the base detector. For the beginning frame $F_0$ of the stream, we duplicate the FPN feature maps as pseudo historical buffers to predict results. This duplication actually means ``no moving'' status and the static results are inconsistent with $F_1$. Fortunately, the performance impact of this situation is minimal, as it is rare.

\subsection{Dual-Flow Perception Module (DFP)}
Given the FPN feature maps of the current frame $F_{t}$ and the historical frame $F_{t-1}$, we assume that the learned features should have two important features for predicting the next frame. One is to use trend features to capture the state of motion and estimate the magnitude of the motion. The other is a basic feature for the detector to localize and classify the corresponding objects.

Therefore, We design a Dual-Flow Perception (DFP) module to encode the expected features with the dynamic flow and static flow, as seen in Fig.~\ref{fig:train}. Dynamic flow fuses the FPN feature from adjacent frames to learn the moving information. It first employs a shared weight $1\times1$ convolution layer followed by the batchnorm and SiLU ~\cite{swish} to reduce the channel to half numbers for both two FPN features. Then, it simply concatenates these two reduced features to generate the dynamic features. We have studied several other fusing operations like add, non-local block~\cite{nonlocal}, STN~\cite{stn} based on squeeze-and-excitation network~\cite{senet}, and correlation layer~\cite{flownet}, where concatenation shows the best efficiency and performance. As for static flow, we reasonably add the original feature of the current frame through a residual connection. In the later experiments, we find the static flow not only provides the basic information for detection but also improves the predicting robustness across different moving speeds of the driving vehicle.

\begin{table*}[t]
\caption{Ablation experiments for building a \emph{simple} and \emph{strong} pipeline. We employ a basic YOLOX-L detector as the baseline for all experiments. Our final designs are marked in \colorbox{baselinecolor}{gray}.}
\vspace{-.2em}
\centering
\subfloat[
\textbf{Sample assignment}. Comparisons on different sample assignment.
\label{tab:sample}
]{
\scalebox{0.8}{
\begin{tabular}{l|c|cc}
\multicolumn{4}{c}{~}\\
\multicolumn{4}{c}{~}\\
\toprule
Method & sAP & $\rm AP_{50}$ & $\rm AP_{75}$\\
\midrule
Baseline & 31.2  & 54.8 & 29.5\\
Latest & 32.8 \textcolor{tablegreen}{\small (\textbf{+1.6)}} & 54.0 & 32.9\\
\baseline{Next} & \baseline{34.2 \textcolor{tablegreen}{\small (\textbf{+3.0)}}} & \baseline{54.6} & \baseline{34.9}\\
Join & 32.3 \textcolor{tablegreen}{\small (\textbf{+1.1)}} & 53.1 & 32.6\\
\bottomrule
\end{tabular}
} 
}
\centering
\hspace{1em}
\subfloat[
\textbf{Prediction task}. Comparisons on different types of prediction tasks. $\dagger$ means auxiliary supervision manner.
\label{tab:Supervision}
]{
\scalebox{0.8}{
\begin{tabular}{l|c|cc}
\toprule
Method & sAP & $\rm AP_{50}$ & $\rm AP_{75}$\\
\midrule
Baseline & 31.2  & 54.8 & 29.5\\
Current offsets & 31.0 \textcolor{tablered}{\small (\textbf{-0.2})} & 52.2 & 30.7\\
Future offsets & 30.3 \textcolor{tablered}{\small (\textbf{-0.9})} & 51.1 & 29.6\\
Current offsets $\dagger$ & 33.6 \textcolor{tablegreen}{\small (\textbf{+2.4})} & 54.1 & 33.3\\
Future offsets $\dagger$ & 33.0 \textcolor{tablegreen}{\small (\textbf{+1.8})} & 54.0 & 32.4\\
\baseline{Next}  & \baseline{34.2 \textcolor{tablegreen}{\small (\textbf{+3.0})}} & \baseline{54.6} & \baseline{34.9}\\
\bottomrule
\end{tabular}
} 
}
\centering
\hspace{1em}
\subfloat[
\textbf{Fusion feature}. Comparisons on three different patterns of features to fuse. $\dagger$ means fusing the input and output features of FPN.
\label{tab:Features}
]{
\scalebox{0.8}{
\begin{tabular}{l|c|ccc}
\multicolumn{4}{c}{~}\\
\toprule
Method & sAP & $\rm AP_{50}$ & $\rm AP_{75}$ & Latency\\
\midrule
Baseline & 31.2  & 54.8 & 29.5 & 18.23 ms\\
Input & 30.3 \textcolor{tablered}{\small (\textbf{-0.9)}} & 50.5 & 29.2 & 18.33 ms \\
Backbone & 30.5 \textcolor{tablered}{\small (\textbf{-0.7)}} & 50.5 & 30.5 & 18.76 ms\\
\baseline{FPN} & \baseline{34.2 \textcolor{tablegreen}{\small (\textbf{+3.0)}}} & \baseline{54.6} & \baseline{34.9} & 18.98 ms\\
FPN $\dagger$  & 33.7 \textcolor{tablegreen}{\small (\textbf{+2.5)}} & 54.3 & 34.7 & 19.42 ms\\
\bottomrule
\end{tabular}
} 
}
\centering
\hspace{1em}
\subfloat[
\textbf{Fusion operator}. Comparisons on different fusion operations. $\dagger$ means concatenating extra correlation features.
\label{tab:operator}
]{
\scalebox{0.8}{
\begin{tabular}{l|c|ccc}
\toprule
Operation & sAP & $\rm AP_{50}$ & $\rm AP_{75}$ & Latency\\
\midrule
Baseline & 31.2  & 54.8 & 29.5 & 18.23 ms\\
Add & 30.8 \textcolor{tablered}{\small (\textbf{-0.4)}} & 54.8 & 29.6 & 26.11 ms\\
STN & 34.0 \textcolor{tablegreen}{\small (\textbf{+2.8)}} & 55.8 & 32.9 & 24.32 ms\\
Correlation  & 32.8 \textcolor{tablegreen}{\small (\textbf{+1.6)}} & 53.9 & 32.9 & 25.22 ms\\
\baseline{Concatenation} & \baseline{34.2 \textcolor{tablegreen}{\small (\textbf{+3.0)}}} & \baseline{54.6} & \baseline{34.9} & 18.98 ms\\
Concatenation $\dagger$ & 33.4 \textcolor{tablegreen}{\small (\textbf{+2.2)}} & 53.6 & 34.2 & 25.78 ms\\
\bottomrule
\end{tabular}
} 
}
\centering
\hspace{1em}
\subfloat[
\textbf{Pre-training policy}. Comparisons on different types of prediction tasks. $\dagger$ and $\dagger$ pre-training with flip and Mosaic respectively.
\label{tab:early}
]{
\scalebox{0.8}{
\begin{tabular}{l|c|c|cc}
\multicolumn{4}{c}{~}\\
\toprule
Method & Off AP & sAP & $\rm AP_{50}$ & $\rm AP_{75}$\\
\midrule
Baseline & - & 31.2  & 54.8 & 29.5\\
\baseline{None} & \baseline{-} & \baseline{34.2 \textcolor{tablegreen}{\small (\textbf{+3.0})}} & \baseline{54.6} & \baseline{34.9}\\
Argoverse & 38.4 & 34.2 \textcolor{tablegreen}{\small (\textbf{+3.0})} & 55.1 & 34.8\\
Argoverse $\dagger$ & 39.7 & 33.9 \textcolor{tablegreen}{\small (\textbf{+2.7})} & 54.9 & 34.9\\
Argoverse $\ddagger$ & 39.1 &  32.7 \textcolor{tablegreen}{\small (\textbf{+1.5})} & 50.7 & 34.1\\
\bottomrule
\end{tabular}
} 
}
\centering
\hspace{1em}
\subfloat[
\textbf{Transfer data augmentation}. Comparisons on different types of prediction tasks.
\label{tab:aug}
]{
\scalebox{0.8}{
\begin{tabular}{l|c|cc}
\multicolumn{4}{c}{~}\\
\multicolumn{4}{c}{~}\\
\toprule
Method & sAP & $\rm AP_{50}$ & $\rm AP_{75}$\\
\midrule
No Aug  & 34.2 \textcolor{tablegreen}{\small (\textbf{+3.0})} & 54.6 & 34.9\\
\baseline{Flip} & \baseline{35.0 \textcolor{tablegreen}{\small (\textbf{+3.8})}} & \baseline{55.4} & \baseline{35.9}\\
hsv & 33.2 \textcolor{tablegreen}{\small (\textbf{+2.0})} & 55.3 & 33.7\\
Mosaic & 29.1 \textcolor{tablered}{\small (\textbf{-1.9})} & 50.2 & 27.4\\
\bottomrule
\end{tabular}
} 
}
\centering

\label{tab:ablations}
\end{table*}

\subsection{Trend-Aware Loss (TAL)}
We notice an important fact in streaming, in which the moving speed of each object within one frame is quite different. The variant trends come from many aspects: different sizes and moving states of their own, occlusions, or the different topological distances. 

Motivated by the observations, we introduce a Trend-Aware Loss (TAL) which adopts adaptive weight for each object according to its moving trend. Generally, we pay more attention to the fast-moving objects as they are more difficult to predict the future states. To quantitatively measure the moving speed, we introduce a trend factor for each object. We calculate an IoU (Intersection over Union) matrix between the ground truth boxes of $F_{t+1}$ and $F_{t}$ and then conduct the max operation on the dimension of $F_{t}$ to get the matching IoU of the corresponding objects between two frames. The small value of this matching IoU means the fast-moving speed of the object and vice versa. 
If a new object comes in $F_{t+1}$, there is no box to match it and its matching IoU is much smaller than usual. We set a threshold $\tau$ to handle this situation and formulate the final trend factor $\omega_i$ for each object in $F_{t+1}$ as:

\begin{equation}
    \label{eq:eq1}
    mIoU_i = \max_j (\{IoU(box^{t+1}_i,box^{t}_{j})\})
\end{equation}

\begin{equation}
    \label{eq:eq2}
    \omega_i = \left\{ {\begin{array}{*{20}{c}}
       {1/mIoU_i  \qquad mIoU_i \ge \tau }
    \\
       {1/\nu  \qquad \qquad mIoU_i < \tau }
\end{array}}, \right.
\end{equation}
where $\max_j$ represents the max operation among boxes in $F_t$, $\nu$ is a constant weight for the new coming objects. We set $\nu$ as 1.6 (bigger than 1) to reduce the attention according to hyper-parameters grid searching.

Note that simply applying the weight to the loss of each object will
change the magnitude of the total losses. This may disturb the balance between the loss of positive and negative samples and decrease the detection performance. Inspired by ~\cite{ggiou,iou_balanced}, we normalize $\omega_{i}$ to $\hat{\omega}_{i}$ intending to keep the sum of total loss unchanged: 

\begin{equation}
    \label{eq:eq3}
    \hat{\omega}_{i} = \omega _i\cdot\frac{{\sum_{i = 1}^N {{\mathcal{L}^{reg}_i}} }}{{\sum_{i = 1}^N {{\omega_i}}{\mathcal{L}^{reg}_i}}},
\end{equation}
where $\mathcal{L}^{reg}_i$ indicates the regression loss of object $i$. Next, we re-weight the regression loss of each object with $\hat{\omega}_{i}$ and the total loss is exhibited as:  

\begin{equation}
    \label{eq:eq4}
    {\mathcal{L}_{total}} = \sum\limits_{i \in positive}\hat{\omega}_{i}\mathcal{L}^{reg}_i  + {\mathcal{L}_{cls}} + {\mathcal{L}_{obj}}.
\end{equation}

\subsection{Velocity-awared streaming AP (VsAP)}
In a realistic driving scene, a vehicle may keep at a standstill, drive slowly and fast. An excellent algorithm ought to be capable of dealing with any traffic case at any driving velocities and even at a standstill. To dig the issue, it needs to rebuild a dataset that is collected at different driving velocities. The collecting process is time-consuming and labor-intensive so we contract and imitate different velocities using off-the-shelf datasets such as Argoverse-HD ~\cite{argoverse,streamer}. We sample two frames at any \emph{intervals} to simulate any velocities. The larger the interval is, the faster the vehicle drives. We copy two identical frames to simulate static. Technologically, we mark the triplet as $\{(F_{t-1}^{mx}$, $F_{t}^{mx}$, $G_{t+1}^{mx})\}_{t=1}^{n_{t}}$, where $n_{t}$ is the total sample number and $mx$ is the m times velocity.

To evaluate the comprehensive performance of different velocities on streaming perception setting, we propose VsAP which simply finds the average value of sAP at various velocity settings. The calculating process is exhibited as:
\begin{equation}
    \label{eq:eq5}
    {\rm VsAP} = \sum\limits_{i \in M}{\rm sAP}^M, M\in[0,6].
\end{equation}

To tackle the more practical issue, we propose to train our detector by randomly sampling triple data at random velocity. The training framework can guide the model to perceive different velocities well (performing better in VsAP). However, the policy may fail to be compatible with TAL, as it will endow higher speed objects with larger loss weight and further bring about imbalance concentration among different velocity settings. In addition, the matching policy is not suitable as the driving speed increases. It may match an error object, leading to unsuccessful training. Facing this dilemma, we employ \emph{actually adjacent frames (30 FPS)} to calculate the corresponding IoU, so that all velocity triples can be trained at an identity loss scale.

\section{Experiments}
\label{sec:exp}


\subsection{Settings}

\paragraph{Datasets} We conduct the experiments on video autonomous driving dataset Argoverse-HD ~\cite{argoverse,streamer} (High-frame-rate Detection),  which contains diverse urban outdoor scenes from two US cities. It has multiple sensors and high frame-rate sensor data (30 FPS). Following ~\cite{streamer}, we only use the center RGB camera and the detection annotations provided by ~\cite{streamer}. We also follow the train/val split in ~\cite{streamer}, where the validation set contains 24 videos with a total of 15k frames.

\paragraph{Evaluation metrics} We use sAP ~\cite{streamer} to evaluate all experiments. sAP is a metric for streaming perception. It simultaneously considers latency and accuracy. Similar to MS COCO metric ~\cite{coco}, it evaluates average mAP over IoU (Intersection-over-Union) thresholds from 0.5 to 0.95 as well as $\rm AP_{s}$, $\rm AP_{m}$, $\rm AP_{l}$ for small, medium and large object. In addition, we report our proposed VsAP for evaluating the comprehensive performance at different velocity settings.

\paragraph{Implementation details} If not specified, we use YOLOX-L ~\cite{yolox} as our default detector and conduct experiment on 1x velocity setting. All of our experiments are fine-tuned from the COCO pre-trained model by 15 epochs. We set batch size at 32 on 8 GTX 2080ti GPUs. We use stochastic gradient descent (SGD) for training. We adopt a learning rate of $0.001 \times BatchSize / 64$ (linear scaling~\cite{linear}) and the cosine schedule with a warm-up strategy for 1 epoch. The weight decay is 0.0005 and the SGD momentum is 0.9. The base input size of the image is $600 \times 960$ while the long side evenly ranges from 800 to 1120 with 16 strides. We only use horizontal flip and do not use other data augmentations (such as Mosaic~\cite{yolo5}, Mixup~\cite{mixup}, hsv etc.) since the feeding adjacent frames need to be aligned. For inference, we keep the input size at $600 \times 960$ and measure the processing time on a Tesla V100 GPU. 

\begin{table}[!htb]
\caption{The effect of the proposed pipeline, DFP, and TAL. 'Off AP' means the corresponding AP using the base detector on the offline setting. 'Pipe.' denotes the proposed pipeline, marked in \colorbox{baselinecolor}{gray}, while '$\uparrow$' indicates the corresponding improvements. '\textcolor{tablegreen}{($\cdot$)}' indicates the relative improvements based on the strong pipeline.} 
\centering
\scalebox{0.9}{
\begin{tabular}{@{}c|ccc|c|c|cc@{}}
\toprule
Model & Pipe. & DFP & TAL & Off AP & sAP & $\rm AP_{50}$ & $\rm AP_{75}$\\
\midrule
\multirow{5}*{YOLOX-S} & & & & \multirow{5}*{31.9} & 26.4 & 48.0 & 24.7\\
~ & \checkmark & & & ~ & \baseline{$\rm 28.1_{\ \uparrow \ 1.7}$}  & \baseline{49.1} & \baseline{27.7}\\
~ &\checkmark &\checkmark & & ~ & 28.9 \textcolor{tablegreen}{\small (\textbf{+0.8)}} & 48.7 & 29.8\\
~ &\checkmark & &\checkmark & ~ & 28.7 \textcolor{tablegreen}{\small (\textbf{+0.6)}} & 48.8 & 29.3\\
~ &\checkmark &\checkmark &\checkmark & ~ & 29.5 \textcolor{tablegreen}{\small (\textbf{+1.4)}} & 50.3 & 29.8\\
\midrule

\multirow{5}*{YOLOX-M} & & & &\multirow{5}*{36.5} & 30.1 & 52.9 & 29.0\\
~ & \checkmark & & & ~ & \baseline{$\rm 32.0_{\ \uparrow \ 1.9}$} & \baseline{52.8} & \baseline{32.5}\\
~ &\checkmark &\checkmark & & ~ & 33.1 \textcolor{tablegreen}{\small (\textbf{+1.1)}} & 54.5 & 33.2\\
~ &\checkmark & &\checkmark & ~ & 32.5 \textcolor{tablegreen}{\small (\textbf{+0.5)}} & 53.1 & 33.0\\
~ &\checkmark &\checkmark &\checkmark & ~ & 33.7 \textcolor{tablegreen}{\small (\textbf{+1.7)}} & 54.5 & 34.0\\

\midrule
\multirow{5}*{YOLOX-L} & & & &\multirow{5}*{39.3} & 32.2 & 55.6 & 30.6\\
~ & \checkmark & & & ~ & \baseline{$\rm 35.0_{\ \uparrow \ 2.8}$} & \baseline{55.4} & \baseline{35.9}\\
~ &\checkmark &\checkmark & &  ~ & 35.7 \textcolor{tablegreen}{\small (\textbf{+0.7)}} & 56.4 & 36.4\\
~ &\checkmark & &\checkmark & ~ & 36.0 \textcolor{tablegreen}{\small (\textbf{+1.0)}} & 57.9 & 36.5\\
~ &\checkmark &\checkmark &\checkmark &~ & 36.9 \textcolor{tablegreen}{\small (\textbf{+1.9)}} & 58.1 & 37.5\\
\bottomrule

\end{tabular}
} 
\label{tab:table3}
\end{table}

\subsection{Ablations for Pipeline}
We conduct ablation studies for the pipeline design on six crucial components: the sample assigning policy the task of prediction, the feature used for fusion, the operation of fusion, the pre-training policy, and the data augmentations for transferring. We employ a basic YOLOX-L detector as the baseline for all experiments and keep the other five components unchanged when ablating one. With these analyses, we show the effectiveness and superiority of our final designs.  

\paragraph{Sample assignment} We study the region of anchor points for assign positive and negative samples, which is a key issue on one-stage detector. Results in Tab.~\ref{tab:sample} shows that specify the anchor (point) region based on the future frame achieve best improvement. It reveals that the assigning manner may impose implicit supervision and reduce the difficulty of regression task.

\paragraph{Prediction task}
\label{exp:pred}
We compare the five types of prediction tasks mentioned in Sec.~\ref{sec:3.2}. 
As shown in Tab.~\ref{tab:Supervision}, indirectly predicting offsets that from latest to current frame (or from current to future frame) with current bounding boxes gets even worse performance than the baseline. In contrast, directly anticipating future results achieves significant improvement (+3.0 AP). Based on directly prediction framework, both type of offsets used to conduct weak supervision lead to performance drop. This demonstrates the supremacy of directly forecasting the results of the next frame. Offsets play a reverse role and may implicate some noise of numerical instability.

\paragraph{Fusion feature} Integrating the current and previous information is the key for the streaming task. For a general detector, there are three different patterns of features to integrate: input, backbone, and FPN pattern respectively. Technically, the input pattern directly concatenates two adjacent frames together and adjusts the input channel of the first layer. The backbone and FPN patterns employ a $1\times1$ convolution followed by batch normalization and SiLU to reduce half channels for each frame and then concatenate them together. We also explore to aggregate the features of FPN and backbone using the same fusing style. As shown in Tab.~\ref{tab:Features}. The results of the input and backbone pattern decrease the performance by 0.9 and 0.7 AP. By contrast, the FPN pattern significantly boosts 3.0 AP, turning into the best choice. Additionally, embedding backbone features into the final enhancing node bring about performance drop (0.5 AP). These results indicate that the fusing FPN feature may get a better trade-off between capturing the motion and detecting the objects. Early fusion may fail to utilize stronger features or disrupt the pre-trained network parameters.

\begin{table}[t]
\caption{Grid search of $\tau$ and $\nu$ in Eq.~\ref{eq:eq2} for TAL.}
\vspace{-.2em}
\centering
\subfloat[
\textbf{$\nu$ is larger than 1}.
\label{tab:table4}
]{
\scalebox{0.9}{
\begin{tabular}{c|c|c|ccccc}
\toprule
$\tau$ & $\nu$ & sAP & $\rm AP_{50}$ & $\rm AP_{75}$ & $\rm AP_{s}$ & $\rm AP_{m}$ & $\rm AP_{l}$\\
\midrule
0.4 & 1.5 & 36.7 & 57.4 & 37.3 & 14.5 & 37.0 & 64.3\\
0.4 & 1.6 & 36.5 & 56.9 & 37.4 & 14.2 & 36.9 & 63.2 \\
0.4 & 1.7 & 36.8 & \textbf{58.3} & 37.1 & 14.6 & 37.4 & 61.5\\
0.5 & 1.5 & 36.7 & 57.2 & 37.2 & 14.7 & 36.7 & 63.9\\
0.5 & 1.6 & \textbf{36.9} & 58.1 & \textbf{37.5} & 14.8 & \textbf{37.5} & \textbf{64.2}\\
0.5 & 1.7 & 36.6 & 57.7 & 37.4 & 14.0 & 37.3 & 62.8\\
0.6 & 1.5 & 36.8 & 57.9 & 37.3 & \textbf{14.9} & 37.0 & 63.3\\
0.6 & 1.6 & 36.4 & 57.3 & 37.0 & 14.5 & 37.1 & 63.7\\
0.6 & 1.7 & 36.2 & 57.8 & 36.5 & 14.2 & 37.1 & 63.2\\
\bottomrule
\end{tabular}
} 
}
\\
\centering
\hspace{2em}
\subfloat[
\textbf{$\nu$ is small than 1}.
\label{tab:small1}
]{
\scalebox{0.9}{
\begin{tabular}{@{}cccccccccc@{}}
\toprule
$\nu$ & 1.0 & 0.9 & 0.8 & 0.7 & 0.8 & 0.7\\
\midrule
sAP  & 36.4 & 36.2 & 36.2 & 35.8  & 35.6 & 35.5\\
\bottomrule
\end{tabular}
} 
}
\centering

\label{tab:tal}
\end{table}



\begin{table}[t]
\caption{Comparison results on different tasks for perceiving trend. 'LOC', 'OBJ', and 'CLS' represent boxes regression, objectness prediction, and classification respectively.} 
\centering
\scalebox{0.85}{
\begin{tabular}{ccc|c|ccccc}
\toprule
 LOC & OBJ & CLS & $\rm mAP$ & $\rm AP_{50}$ & $\rm AP_{75}$ & $\rm AP_{s}$ & $\rm AP_{m}$ & $\rm AP_{l}$\\
\midrule
\checkmark & & & \textbf{36.9} & \textbf{58.1} & \textbf{37.5}  & \textbf{14.8} & \textbf{37.5} & \textbf{64.2}\\
\checkmark & \checkmark &  & 35.7 & 57.3 & 35.7 & 14.0 & 36.7 & 61.9\\
\checkmark & & \checkmark & 36.6 & 56.8 & 37.3 & 14.7 & 36.8 & 63.1\\
\checkmark & \checkmark & \checkmark & 36.3 & 57.5 & 36.3 & 14.4 & 37.0 & 63.4\\
\bottomrule
\end{tabular}} 
\label{tab:perceving_task}
\end{table}

\begin{figure*}[t]
\setlength{\abovecaptionskip}{0pt}
\begin{center}
\includegraphics[width=0.9\linewidth]{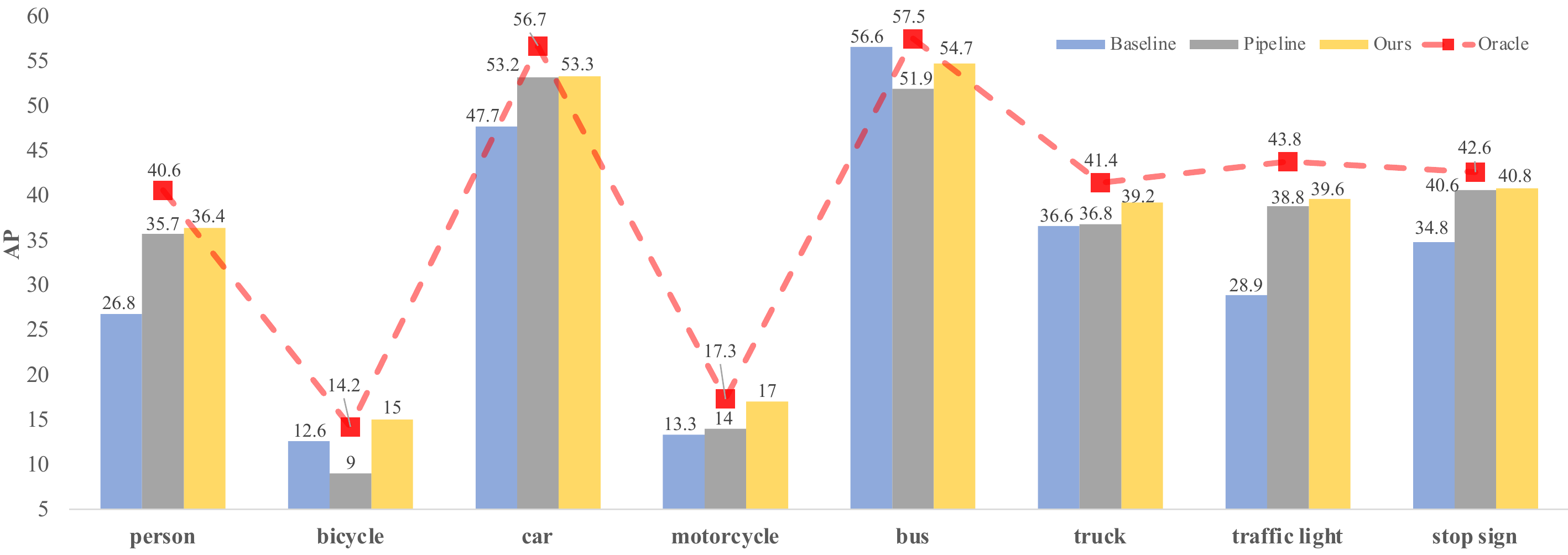}
\end{center}
\caption{Detection results respective to 8 corresponding categories on Argoverse-HD dataset. }
\vspace{-0.4cm}
\label{fig:cls}
\end{figure*}

\paragraph{Fusion operation} We also explore the fusion operation for FPN features. We seek several regular operators (\emph{i.e.}, element-wise add and concatenation) and advanced ones (\emph{i.e.}, spatial transformer network ~\cite{stn} (STN)\footnote{To implement STN for variable inputs, we adopt a SE ~\cite{senet} block to calculate the transformation parameters instead of using flatten operation and fully connected layers in the original STN.}, non-local network ~\cite{nonlocal} (NL)\footnote{For NL, we use the current feature to calculate the values and queries and use the previous feature to generate keys, and then feed to original NL module.} and correlation layer~\cite{flownet}. Tab.~\ref{tab:operator} displays the performance among these operations. We can see that the element-wise add operation drops performance by 0.4 AP while other ones achieve considerable gains. We suppose that adding element-wise values may break down the relative information between two frames and fail to learn trending information. And among effective operations, concatenation is prominent because of its light parameters and high inference speed.

\paragraph{Early fine-tuning} We further to study the effect of early fine-tuning on offline setting, aiming to exploring whether it helps the model learn the knowledge of the target domain. To this end, we early fine-tuning our detector with three types of augmentations (inheriting from YOLOX~\cite{yolox}) and compares with directly fine-tuning by online manner. Results from Tab.~\ref{tab:early} reveals that early fine-tuning achieve the same performance with directly fine-tuning (34.2 vs. 34.2). Early fine-tuning with augmentation achieves better performance on offline setting but having a negative impact on final fine-tuning. Therefore, we choose to directly conduct future prediction task, which do not need to execute redundant multi-stage training.

\paragraph{Data augmentation} Next, we continue to explore how to add data augmentation for future prediction. It is burdensome since unfit augmentation may lead to misalignment between current and previous features and further fail to perceive motion information. We conduct the corresponding transform on input image and annotation to make sure error-free supervision. Flip shows the best performance than other three counterparts. Mosaic brings about performance drop as it actually shrink the object and increase task difficulty. In fact, flip implicitly exposes the model to anticipate different driving directions so that it learn a better generalization.

Based on above ablation results, we can not only establish a stronger baseline, but work with concise training manner, light parameters and high inference speed.

\subsection{Ablations for DFP and TAL}

\paragraph{Effect of DFP and TAL} To validate the effect of DFP and TAL, we conduct extensive experiments on YOLOX detectors with different model sizes. In Tab.~\ref{tab:table3}, ``Pipe.'' denotes our basic pipeline containing basic feature fusion and future prediction training. Compared to the baseline detector, the proposed pipelines have already improved the performance by 1.3 to 3.0 AP across different models. Based on these high-performance baselines, DFP and TAL can boost the accuracy of sAP by $\sim$1.0 AP independently, and their combinations further improve the performance by nearly 2.0 AP. These facts not only demonstrate the effectiveness of DFP and TAL but also indicate that the contributions of the two modules are almost orthogonal. 

Indeed, DFP adopts dynamic flow and static flow to extract the moving state feature and basic detection feature separately and enhances the FPN feature for streaming perception. Meanwhile, TAL employs adaptive weight for each object to predict different trending. We believe the two modules cover different points for streaming perception: architecture and optimization. We hope that our simple design of the two modules will lead to future endeavors in these two under-explored aspects.

\paragraph{Grid search of $\tau$ and $\nu$} As depicted in Eq.~\ref{eq:eq2}, the value of $\tau$ acts as a threshold to monitor newly emerging objects while $\nu$ controls the degree of attention on the new objects. Increasing $\tau$ too much may mislabel the fast-moving objects as new ones and decrease their weights while setting $\tau$ too small will lead to missing some new objects. Therefore, The value of $\tau$ needs to be fine-tuned. As for $\nu$, we set it larger than 1.0 so that the model pays less attention to the new-coming objects. We conduct a grid search for the two hyperparameters. Results in Tab.~\ref{tab:table4} shows that $\tau$ and $\nu$ achieve the best performance at 0.5 and 1.6 respectively. Tab.~\ref{tab:small1} supports the relevant statement ($\nu$ should be larger than 1). As $\nu$ is small than 1, it damages model performance, as more attention should not be placed on theoretically unpredictable new objects.

\paragraph{Perceiving task for TAL} Tab.~\ref{tab:perceving_task} compares different tasks for conducting TAL. The comparison results show that only takes regression task into account achieve the best performance. Reweighting objectness prediction loss causes performance degradation while classification counterpart can no further gain. It reveals that adopting relative speed of objects as metric to more concentrate faster objects' regression task promote future perception.

\begin{figure}[!htb]
\begin{center}
\includegraphics[width=\linewidth]{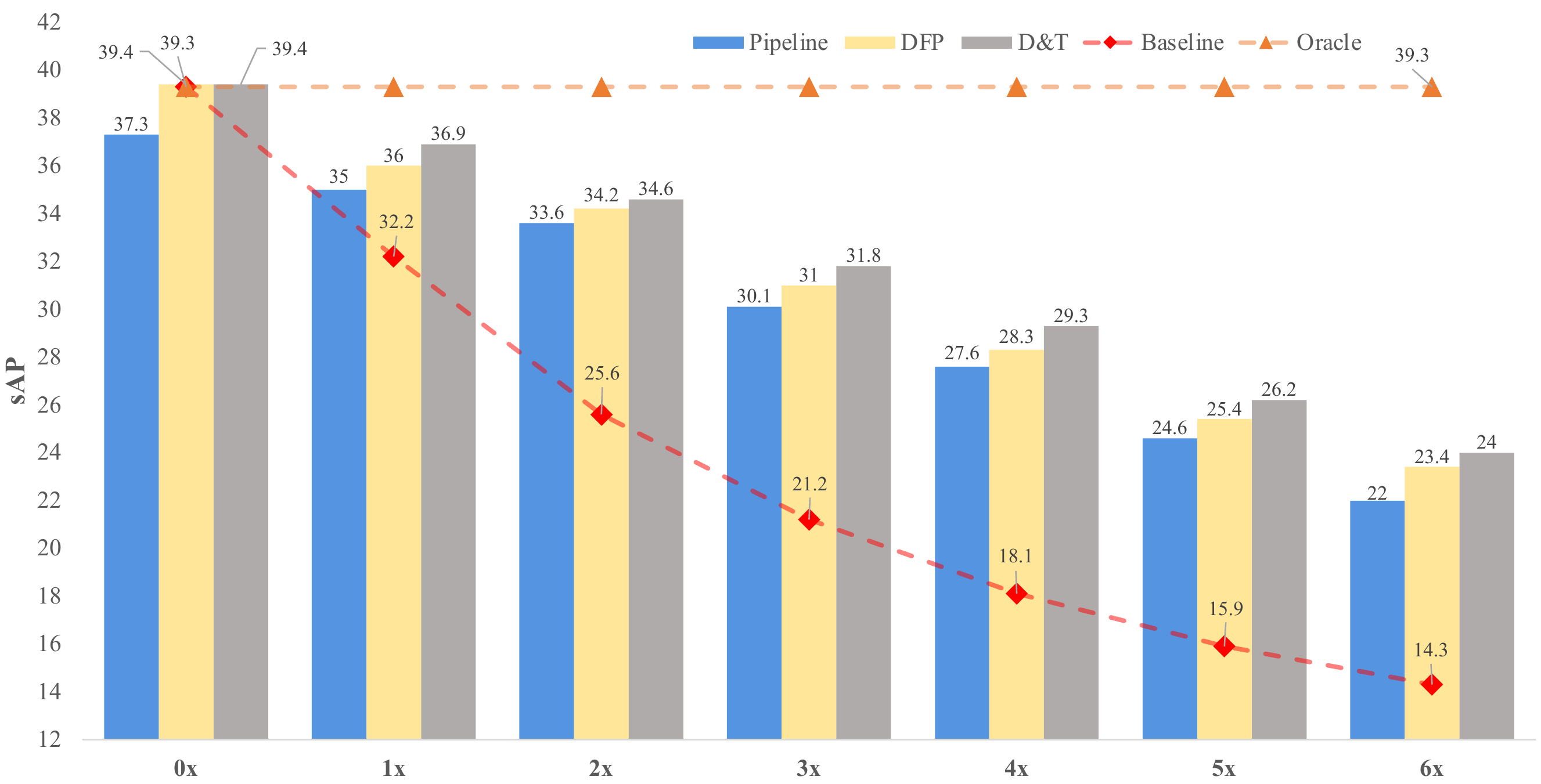}
\caption{Results on different moving velocity settings. 'D\&T' represents DFP and TAL.}
\label{tab:table5}
\end{center}
\end{figure}

\begin{figure}[!htb]
\begin{center}
\includegraphics[width=\linewidth]{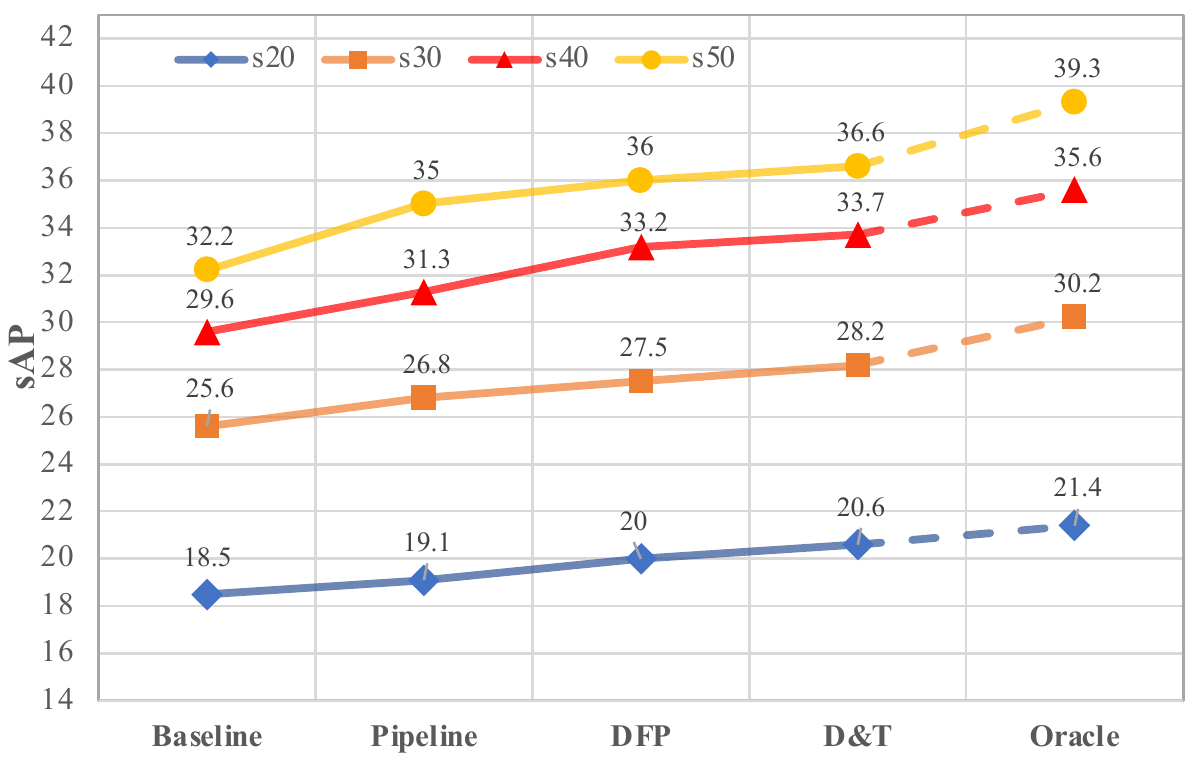}
\caption{Results on different image scales. 'D\&T' represents DFP and TAL. The number after 's' is the input scale (the full resolution is $1200 \times 1920$).}
\label{tab:scales}
\end{center}
\end{figure}

\begin{table}[!htb]
\caption{Comparison results on different forecasting manners.} 
\centering
\scalebox{0.9}{
\begin{tabular}{@{}c|cc|c@{}}
\toprule
Forecasting manner &  $\rm sAP_{1x}$ & $\rm sAP_{2x}$ & Extra Latency\\
\midrule
 Offline Det   &  31.2 & 24.9 & 0 ms\\
KF Forecasting &  35.5 & 31.8 & 3.11 ms\\
 Ours (E2E) &  \textbf{36.1} & \textbf{33.3} & \textbf{0.8 ms}\\
\bottomrule
\end{tabular}
} 
\label{tab:rebuttal_prediction}
\end{table}

\begin{figure*}[!htb]
\centering      
\subfloat[1x velocity setting]
{\includegraphics[width =0.97\linewidth]{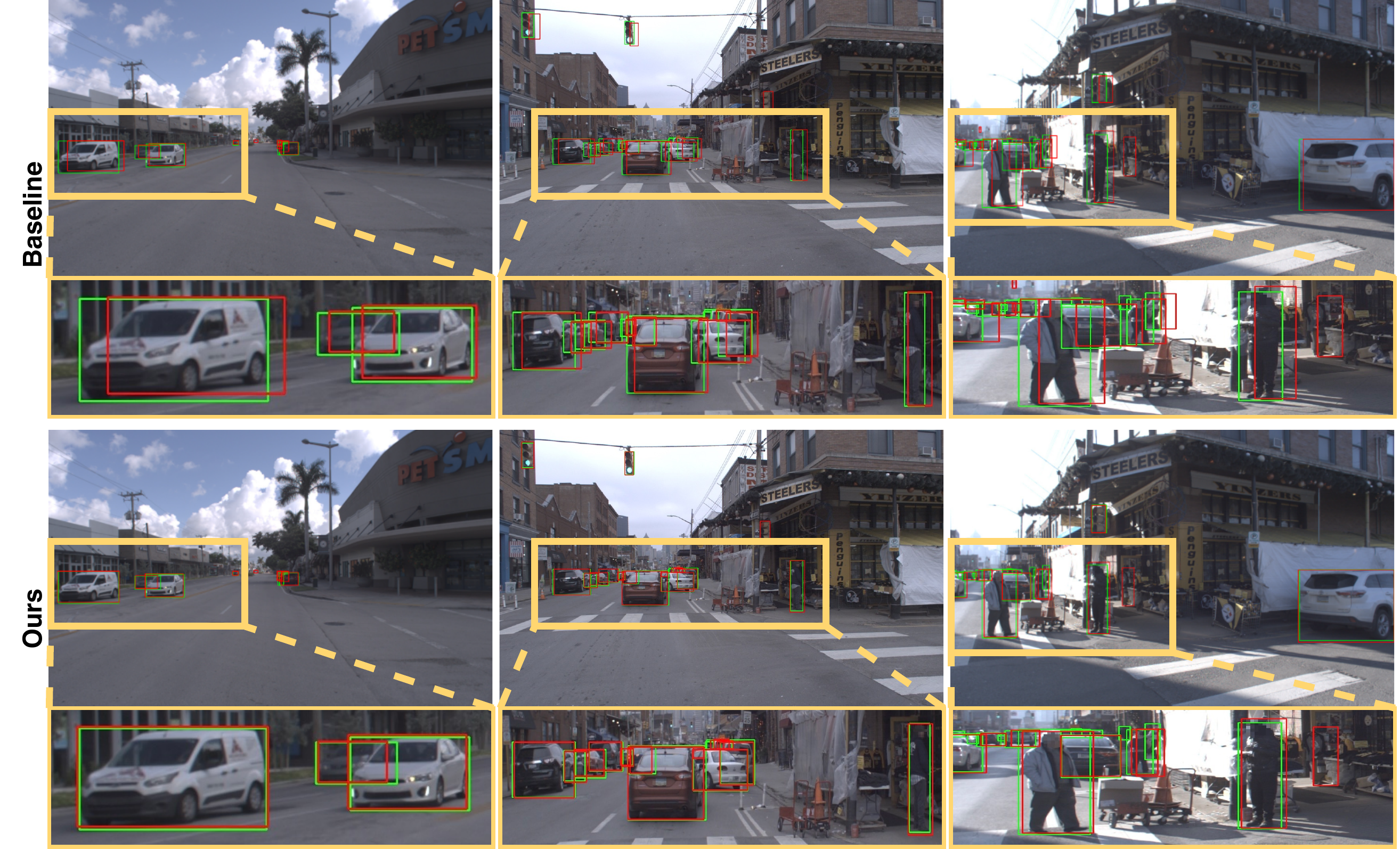}
\label{fig:baseline}}\ \      
\subfloat[3x velocity setting]
{\includegraphics[width = 0.97\linewidth]{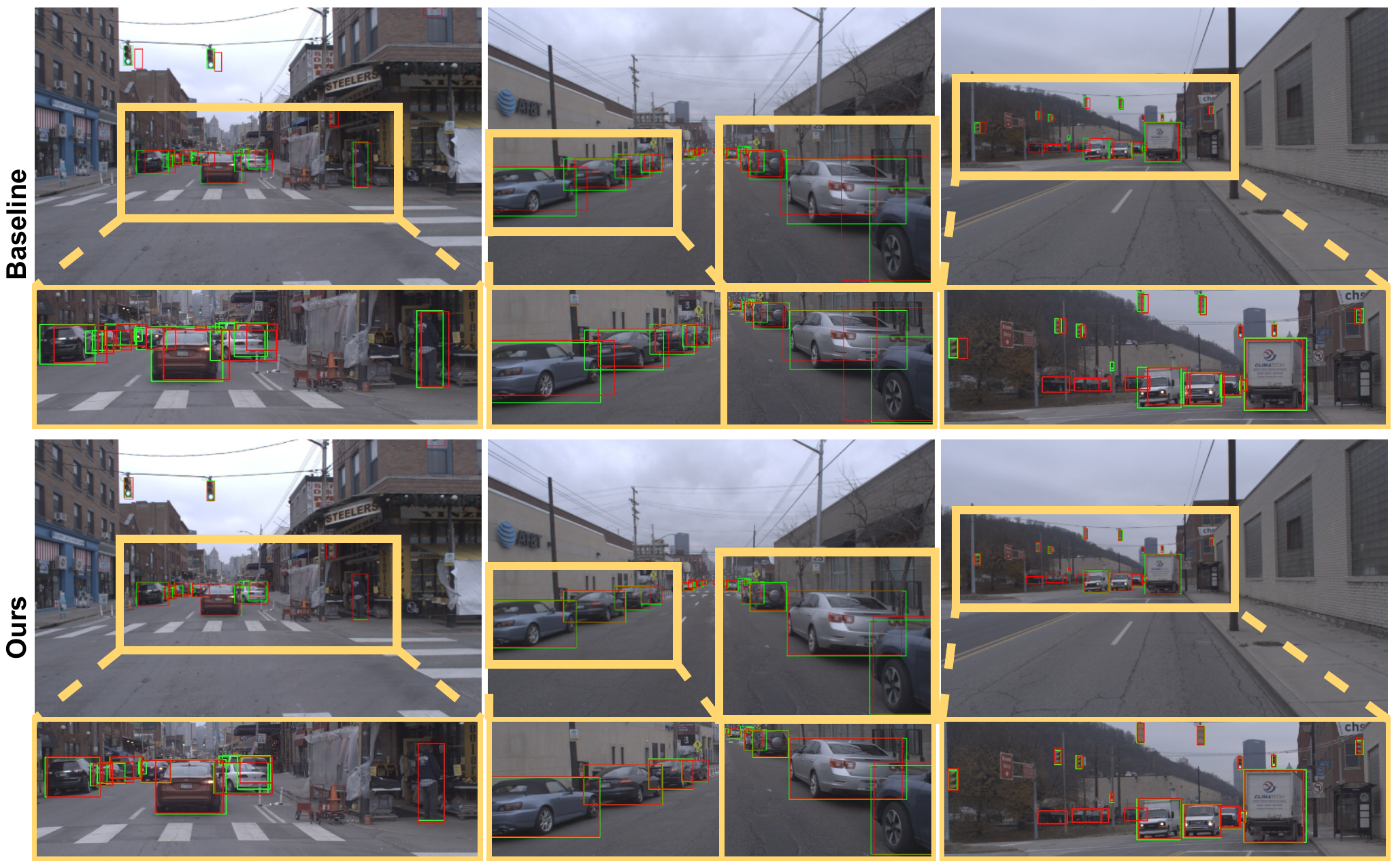}
\label{fig:ours}}\ \      
\caption{Visualization results of the baseline detector and the proposed method. The green boxes represent ground truth boxes, while red ones represent prediction results.}      
\label{fig:visualization}      
\end{figure*}


\begin{table}[!htb]
\caption{Performance comparison with state-of-the-art approaches on Argoverse-HD dataset. Size means the shortest side of input image and the input image resolution is 600$\times$960 for our models. $\dagger$ represents using extra datasets (BDD100K~\cite{yu2020bdd100k}, Cityscapes~\cite{cordts2016cityscapes}, and nuScenes~\cite{caesar2020nuscenes}) for pre-training. $\ddagger$ means using extra dataset and TensorRT.} 
\centering
\resizebox{\linewidth}{!}{ 
\begin{tabular}{@{}c|ccc|ccc@{}}
\toprule
Method & sAP & $\rm AP_{50}$ & $\rm AP_{75}$ & $\rm AP_{s}$ & $\rm AP_{m}$ & $\rm AP_{l}$\\
\midrule
\multicolumn{7}{c}{\textbf{Non-real-time methods}}\\
\midrule
Streamer (size=900)~\cite{streamer} & 18.2 & 35.3 & 16.8 & 4.7 & 14.4 & 34.6\\
Streamer (size=600)~\cite{streamer} & 20.4 & 35.6 & 20.8 & 3.6 & 18.0 & 47.2\\
Streamer + AdaScale~\cite{adascale,adaptivestreamer} & 13.8 & 23.4 & 14.2 & 0.2 & 9.0 & 39.9\\
Adaptive Streamer~\cite{adaptivestreamer} & 21.3 & 37.3 & 21.1 & 4.4 & 18.7 & 47.1\\
\midrule
\multicolumn{7}{c}{\textbf{Real-time methods}}\\
\midrule
1st place $\ddagger$ (size=1440)~\cite{yoloxx} & 40.2 & 68.9 & 39.4 & 21.5 & 42.9 & 53.9\\
2nd place $\ddagger$ (size=1200)~\cite{2nd} & 33.2 & 58.6 & 30.9 & 13.3 & 31.9 & 40.0\\
Ours-S & 29.5 & 50.3 & 29.8 & 11.0 & 30.9 & 51.5\\
Ours-M & 33.7 & 54.5 & 34.0 & 13.2 & 35.3 & 58.7\\
Ours-L & 36.9 & 58.1 & 37.5 & 14.8 & 37.5 & 64.2\\
Ours-L $\dagger$ & 41.1 & 63.6 & 45.2 & 18.0 & 43.4 & 67.1\\
Ours-L $\ddagger$ & 42.3 & 64.5 & 46.4 & 23.9 & 45.7 & 68.1\\
\bottomrule
\end{tabular}
} 

\label{tab:table6}
\end{table}

\begin{figure*}[t]
\begin{center}
\includegraphics[width=0.9\linewidth]{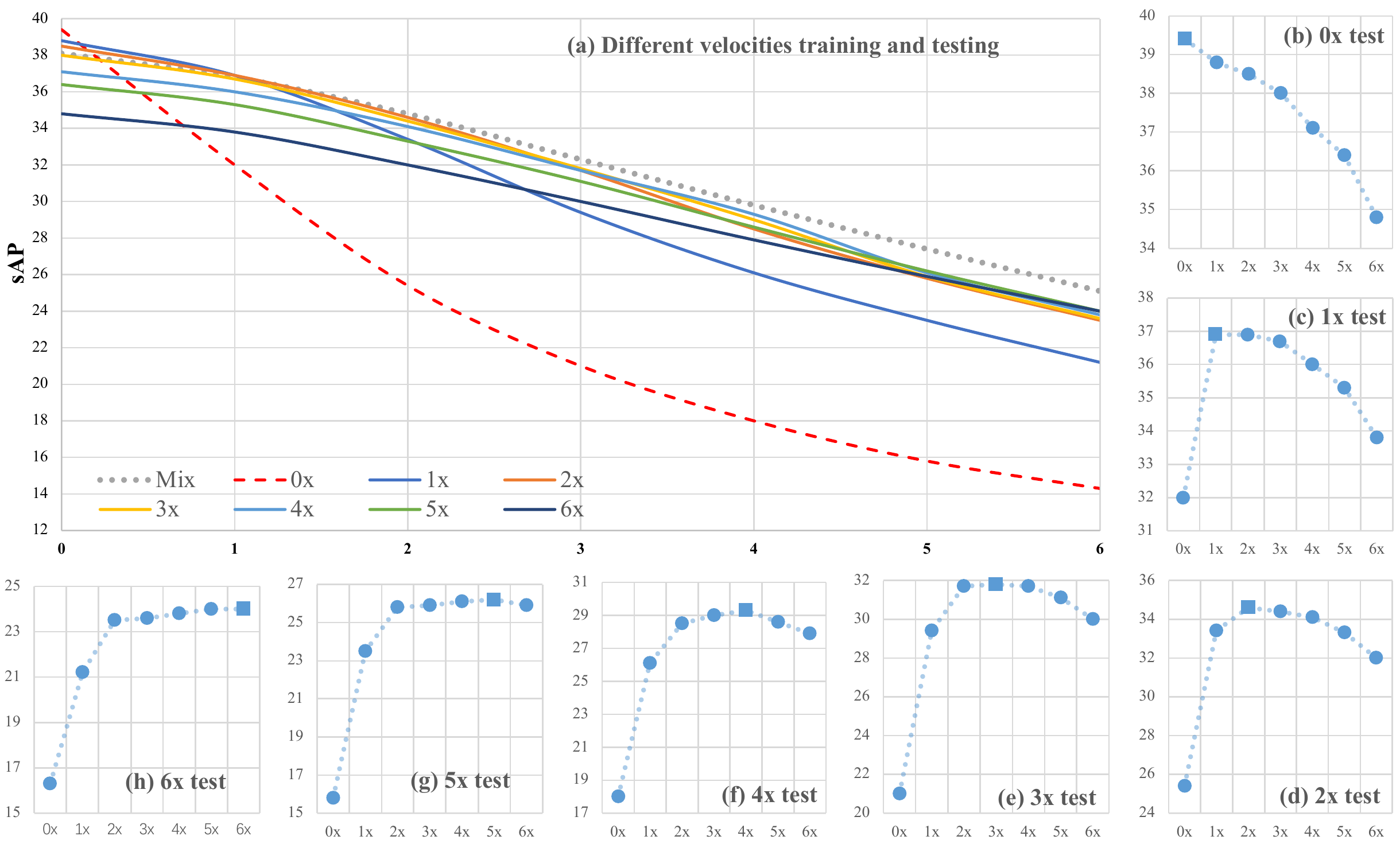}
\caption{Results on different velocities training and testing. Subgraph (a) illustrates training with single velocity and test on different velocities. The stride of input images is simulated correspondingly to velocity during testing. Subgraph (b)-(h) display the performance of training with various velocities and testing on single velocity.}
\label{fig:vsap}
\end{center}
\end{figure*}

\begin{table*}[!htb]
\caption{Comparison results on different velocity training strategies. '*' means advanced TAL with the velocity estimation using actual adjacent frames.} 
\centering
\begin{tabular}{ccccccccccc}
\toprule
Training velocity & Offline & 0x & 1x & 2x & 3x & 4x & 5x & 6x & $\rm Mix$ & $\rm Mix^*$\\
\midrule
Accuracy (VsAP) & 23.8 & 23.7 & 29.9 & 31.36 & 31.34 & 31.16 & 30.7 & 29.77 & 31.02 & \textbf{32.04}\\
\bottomrule
\end{tabular} 
\label{tab:vsap}
\end{table*}

\subsection{Further Analysis}
\paragraph{Performance on different categories} In detail, we compare the performance with respective to different categories. The results are displayed in Fig.~\ref{fig:cls}. By directly utilizing the base detector to detect vehicles, different types of objects behave variously. Small or slender vehicles like person, traffic light, and stop sign show a significantly performance drop, while large ones like bus and trunk are less affected. The phenomenon is reasonable as the shifts of prediction boxes caused by the processing time delay become larger with the scale reducing. For larger objects, the shifts has little effect on the change of IoU. More correspondingly detail visualization results can be see in Fig.~\ref{fig:visualization}. By embedding our proposed DFP and TAL, the performance gaps of different categories are significantly narrowed. It is noting that the detecting ability of some categories like bicycle and motorcycle close to the performance of oracle. 

\paragraph{Robustness at different speeds} Actually, we expect our method not only is capable of alleviating the issue caused by time delay at any speed, but achieve the same results as the offline performance. To this end, we further verify the robustness of our model at different moving speeds of the driving vehicle including static. To simulate the static (0$\times$ speed) and faster speed (2$\times$, \dots, 6$\times$) environments, we re-sample the video frames to build new datasets. For 0$\times$ speed setting, we treat it as a special driving state and re-sample the frames to the triplet $(F_{t}, F_{t}, F_{t})$. It means the previous and current frames have no change and the model should predict the non-changed results.
For N$\times$ speed setting, we re-build the triplet data as  $(F_{t-N}, F_{t}, F_{t+N}), N\in[2,6]$. This indicates the faster moving speed of both the ground truth objects and the driving vehicle.

Results are listed in Fig.~\ref{tab:table5}. For 0$\times$ speed, the predicting results are supposed to be the same as the offline setting. However, if we only adopt the basic pipeline, we can see a significant performance drop (-2.0 AP) compared to the offline, which means the model fails to deduce the static state. By adopting the DFP module into the basic pipeline, we recover this reduction and achieve the 
comparable results as the offline performance. It reveals that DFP, especially the static flow, is a key to extracting the right moving trend and assisting in prediction. It is also worth noting that at 0$\times$ speed, all the weights in TAL are one thus it has no influence. For N$\times$ speed, as the objects move faster, the gap between offline and streaming perception is further expanded. Meanwhile, the improvements from our models, including the basic pipeline, DFP, and TAL, are also enlarged. These robustness results further manifest the superiority of our method. 

\paragraph{Robustness at different scales} We further verify the performance of our method at different image scales. Results are shown in Fig.~\ref{tab:scales}. The strong pipeline with DFP and TAL modules continuously narrows the performance gap between offline and streaming perception settings. It illustrates that our approach and pipeline can (re)act at any input scales, which is prone to embed in different devices.

\paragraph{Comparison with Kalman Filter based forecasting} We follow the implementation of ~\cite{streamer} and report the results in Tab.~\ref{tab:rebuttal_prediction}. For ordinary sAP ($1\times$), our end-to-end method still outperforms the advanced baseline by 0.5 AP. Further, when we simulate and evaluate them with faster moving ($2\times$ speed), our end-to-end model shows more superiority of robustness (33.3 sAP v.s. 31.8 sAP). Besides, our e2e model brings lower extra latency (0.8 ms v.s. 3.1 ms taking the average of 5 tests). 

\paragraph{Visualization results} As shown in Fig.~\ref{fig:visualization}, we present the visualization results by comparing the base detector and our method. For the baseline detector, the predicting bounding boxes encounter severe time lag. The faster the vehicles and pedestrians move, the larger the predictions shift. For small objects like traffic lights, the overlap between predictions and ground truth becomes small and is even non. In contrast, our method alleviates the mismatch and fits accurately between the predicting boxes and moving objects. It further confirms the effectiveness of our method visually.

\subsection{Actual analysis}

To further reveal the more actual scenes with different driving velocities, we conduct extensive experiments to analyse the regular pattern with different velocity training settings. Afterwards, we carry out comparison experiments to show the superiority of mix-velocity training strategy with DFP and TAL.

The Fig.~\ref{fig:vsap} show the interesting results by single velocity training and various velocities testing. It reveals that the closer to the training velocity, the more testing performance gains, which is similar to the performance of different training image scales and testing image scales. The line chart shows that DFP and TAL stably narrow the performance drop by a little performance decrease of static at the expense. As shown in Tab.~\ref{tab:vsap}, we use the VsAP to measure the overall performance of all velocity settings. The fixed single velocity training framework need to search a better trade-off, like 2x achieves a better performance (32.36 VsAP). Different from the time-consuming and unscientific single-velocity training method, we use the mix-velocity training manner, which achieves the best performance (32.04 VsAP) and gains 8.24\% improvement. We also compare the mix training with vanilla and advanced TAL. Results shows that the vanilla achieve a decent effect but is lower than the single velocity style searching manner, which may be caused by the imbalance between loss weights at different speeds. As a comparison, the advanced TAL adopting actually adjacent frames achieves higher gains.

\subsection{Comparison with state-of-the-art}
We compare our method with other state-of-the-art detectors on Argoverse-HD dataset. As shown in Fig.~\ref{tab:table6}, real-time methods show absolute advantages over non-real-time detectors.
We also report the results of the 1st and 2nd place in Streaming Perception Challenge. They involve extra datasets and accelerating tricks which are out of the scope for this work, while our methods get competitive performance and even surpass the accuracy of the 2nd place scheme without any tricks. By adopting extra datasets for pre-training our method surpass the accuracy of the 1st place scheme only using 600 $\times$ 960 resolution. With larger resolution inputs (1200 $\times$ 1920), our method further achieve better performance (42.3 sAP).


\section{Conclusion}
This paper focuses on a streaming perception task that takes the processing latency into account. Under this metric, we reveal the superiority of using a real-time detector with the ability of future prediction for online perception. We further build a real-time detector with Dual-Flow Perception module and Trend-Aware Loss, alleviating the time lag problem in streaming perception. Extensive experiments show that our simple framework achieves superior performance against all state-of-the-art detectors. It also obtains robust results on different speed settings. Further, we synthetically consider the driving velocity and propose the mix-velocity training strategy with advanced TAL to training our detector for adapting to various velocities. We hope that our simple and effective design will motivate future efforts in this practical and challenging perception task.

\ifCLASSOPTIONcaptionsoff
  \newpage
\fi



\bibliographystyle{IEEEtran}
\bibliography{IEEEabrv, cite}

\begin{thebibliography}{10}
\providecommand{\url}[1]{#1}
\csname url@samestyle\endcsname
\providecommand{\newblock}{\relax}
\providecommand{\bibinfo}[2]{#2}
\providecommand{\BIBentrySTDinterwordspacing}{\spaceskip=0pt\relax}
\providecommand{\BIBentryALTinterwordstretchfactor}{4}
\providecommand{\BIBentryALTinterwordspacing}{\spaceskip=\fontdimen2\font plus
\BIBentryALTinterwordstretchfactor\fontdimen3\font minus
  \fontdimen4\font\relax}
\providecommand{\BIBforeignlanguage}[2]{{%
\expandafter\ifx\csname l@#1\endcsname\relax
\typeout{** WARNING: IEEEtran.bst: No hyphenation pattern has been}%
\typeout{** loaded for the language `#1'. Using the pattern for}%
\typeout{** the default language instead.}%
\else
\language=\csname l@#1\endcsname
\fi
#2}}
\providecommand{\BIBdecl}{\relax}
\BIBdecl

\bibitem{yolo1}
J.~Redmon, S.~Divvala, R.~Girshick, and A.~Farhadi, ``You only look once:
  Unified, real-time object detection,'' in \emph{Proceedings of the IEEE
  conference on computer vision and pattern recognition}, 2016, pp. 779--788.

\bibitem{yolo2}
J.~Redmon and A.~Farhadi, ``Yolo9000: better, faster, stronger,'' in
  \emph{Proceedings of the IEEE conference on computer vision and pattern
  recognition}, 2017, pp. 7263--7271.

\bibitem{yolo3}
------, ``Yolov3: An incremental improvement,'' \emph{arXiv preprint
  arXiv:1804.02767}, 2018.

\bibitem{yolo4}
A.~Bochkovskiy, C.-Y. Wang, and H.-Y.~M. Liao, ``Yolov4: Optimal speed and
  accuracy of object detection,'' \emph{arXiv preprint arXiv:2004.10934}, 2020.

\bibitem{yolo5}
glenn jocher, ``yolov5,'' \emph{https://github.com/ultralytics/yolov5}, 2021.

\bibitem{yolox}
Z.~Ge, S.~Liu, F.~Wang, Z.~Li, and J.~Sun, ``Yolox: Exceeding yolo series in
  2021,'' \emph{arXiv preprint arXiv:2107.08430}, 2021.

\bibitem{streamer}
M.~Li, Y.-X. Wang, and D.~Ramanan, ``Towards streaming perception,'' in
  \emph{European Conference on Computer Vision}.\hskip 1em plus 0.5em minus
  0.4em\relax Springer, 2020, pp. 473--488.

\bibitem{htc}
K.~Chen, J.~Pang, J.~Wang, Y.~Xiong, X.~Li, S.~Sun, W.~Feng, Z.~Liu, J.~Shi,
  W.~Ouyang \emph{et~al.}, ``Hybrid task cascade for instance segmentation,''
  in \emph{Proceedings of the IEEE/CVF Conference on Computer Vision and
  Pattern Recognition}, 2019, pp. 4974--4983.

\bibitem{retinanet}
T.-Y. Lin, P.~Goyal, R.~Girshick, K.~He, and P.~Doll{\'a}r, ``Focal loss for
  dense object detection,'' in \emph{Proceedings of the IEEE international
  conference on computer vision}, 2017, pp. 2980--2988.

\bibitem{maskrcnn}
K.~He, G.~Gkioxari, P.~Doll{\'a}r, and R.~Girshick, ``Mask r-cnn,'' in
  \emph{Proceedings of the IEEE international conference on computer vision},
  2017, pp. 2961--2969.

\bibitem{adaptivestreamer}
A.~Ghosh, A.~Nambi, A.~Singh, H.~YVS, and T.~Ganu, ``Adaptive streaming
  perception using deep reinforcement learning,'' \emph{arXiv preprint
  arXiv:2106.05665}, 2021.

\bibitem{fovea}
C.~Thavamani, M.~Li, N.~Cebron, and D.~Ramanan, ``Fovea: Foveated image
  magnification for autonomous navigation,'' in \emph{Proceedings of the
  IEEE/CVF International Conference on Computer Vision}, 2021, pp.
  15\,539--15\,548.

\bibitem{yoloxx}
S.~Zhang, L.~Song, S.~Liu, Z.~Ge, Z.~Li, X.~He, and J.~Sun, ``Workshop on
  autonomous driving at cvpr 2021: Technical report for streaming perception
  challenge,'' \emph{arXiv preprint arXiv:2108.04230}, 2021.

\bibitem{2nd}
Y.~Gu, Q.~Wang, and X.~Qin, ``Real-time streaming perception system for
  autonomous driving,'' \emph{arXiv preprint arXiv:2107.14388}, 2021.

\bibitem{kalman}
R.~E. Kalman, ``A new approach to linear filtering and prediction problems,''
  \emph{Transactions of the ASME–Journal of Basic Engineering 82(Series D)},
  pp. 35--45, 1960.

\bibitem{argoverse}
M.-F. Chang, J.~Lambert, P.~Sangkloy, J.~Singh, S.~Bak, A.~Hartnett, D.~Wang,
  P.~Carr, S.~Lucey, D.~Ramanan \emph{et~al.}, ``Argoverse: 3d tracking and
  forecasting with rich maps,'' in \emph{Proceedings of the IEEE/CVF Conference
  on Computer Vision and Pattern Recognition}, 2019, pp. 8748--8757.

\bibitem{yang2022real}
J.~Yang, S.~Liu, Z.~Li, X.~Li, and J.~Sun, ``Real-time object detection for
  streaming perception,'' \emph{arXiv preprint arXiv:2203.12338}, 2022.

\bibitem{fastrcnn}
R.~Girshick, ``Fast r-cnn,'' in \emph{Proceedings of the IEEE international
  conference on computer vision}, 2015, pp. 1440--1448.

\bibitem{fpn}
T.-Y. Lin, P.~Doll{\'a}r, R.~Girshick, K.~He, B.~Hariharan, and S.~Belongie,
  ``Feature pyramid networks for object detection,'' in \emph{Proceedings of
  the IEEE conference on computer vision and pattern recognition}, 2017, pp.
  2117--2125.

\bibitem{fasterrcnn}
S.~Ren, K.~He, R.~Girshick, and J.~Sun, ``Faster r-cnn: Towards real-time
  object detection with region proposal networks,'' \emph{Advances in neural
  information processing systems}, vol.~28, pp. 91--99, 2015.

\bibitem{ssd}
W.~Liu, D.~Anguelov, D.~Erhan, C.~Szegedy, S.~Reed, C.-Y. Fu, and A.~C. Berg,
  ``Ssd: Single shot multibox detector,'' in \emph{European conference on
  computer vision}.\hskip 1em plus 0.5em minus 0.4em\relax Springer, 2016, pp.
  21--37.

\bibitem{fcos}
Z.~Tian, C.~Shen, H.~Chen, and T.~He, ``Fcos: Fully convolutional one-stage
  object detection,'' in \emph{Proceedings of the IEEE/CVF international
  conference on computer vision}, 2019, pp. 9627--9636.

\bibitem{cspnet}
C.-Y. Wang, H.-Y.~M. Liao, Y.-H. Wu, P.-Y. Chen, J.-W. Hsieh, and I.-H. Yeh,
  ``Cspnet: A new backbone that can enhance learning capability of cnn,'' in
  \emph{Proceedings of the IEEE/CVF conference on computer vision and pattern
  recognition workshops}, 2020, pp. 390--391.

\bibitem{pan}
S.~Liu, L.~Qi, H.~Qin, J.~Shi, and J.~Jia, ``Path aggregation network for
  instance segmentation,'' in \emph{Proceedings of the IEEE conference on
  computer vision and pattern recognition}, 2018, pp. 8759--8768.

\bibitem{mixup}
H.~Zhang, M.~Cisse, Y.~N. Dauphin, and D.~Lopez-Paz, ``mixup: Beyond empirical
  risk minimization,'' \emph{arXiv preprint arXiv:1710.09412}, 2017.

\bibitem{copypaste}
G.~Ghiasi, Y.~Cui, A.~Srinivas, R.~Qian, T.-Y. Lin, E.~D. Cubuk, Q.~V. Le, and
  B.~Zoph, ``Simple copy-paste is a strong data augmentation method for
  instance segmentation,'' in \emph{Proceedings of the IEEE/CVF Conference on
  Computer Vision and Pattern Recognition}, 2021, pp. 2918--2928.

\bibitem{mega}
Y.~Chen, Y.~Cao, H.~Hu, and L.~Wang, ``Memory enhanced global-local aggregation
  for video object detection,'' in \emph{Proceedings of the IEEE/CVF Conference
  on Computer Vision and Pattern Recognition}, 2020, pp. 10\,337--10\,346.

\bibitem{rdn}
J.~Deng, Y.~Pan, T.~Yao, W.~Zhou, H.~Li, and T.~Mei, ``Relation distillation
  networks for video object detection,'' in \emph{Proceedings of the IEEE/CVF
  International Conference on Computer Vision}, 2019, pp. 7023--7032.

\bibitem{rnl}
H.~Hu, J.~Gu, Z.~Zhang, J.~Dai, and Y.~Wei, ``Relation networks for object
  detection,'' in \emph{Proceedings of the IEEE conference on computer vision
  and pattern recognition}, 2018, pp. 3588--3597.

\bibitem{dff}
X.~Zhu, Y.~Xiong, J.~Dai, L.~Yuan, and Y.~Wei, ``Deep feature flow for video
  recognition,'' in \emph{Proceedings of the IEEE conference on computer vision
  and pattern recognition}, 2017, pp. 2349--2358.

\bibitem{fgfa}
X.~Zhu, Y.~Wang, J.~Dai, L.~Yuan, and Y.~Wei, ``Flow-guided feature aggregation
  for video object detection,'' in \emph{Proceedings of the IEEE International
  Conference on Computer Vision}, 2017, pp. 408--417.

\bibitem{lstm1}
Q.~Zhou, X.~Liang, K.~Gong, and L.~Lin, ``Adaptive temporal encoding network
  for video instance-level human parsing,'' in \emph{Proceedings of the 26th
  ACM international conference on Multimedia}, 2018, pp. 1527--1535.

\bibitem{lstm2}
Z.~Pi, H.~Qin, C.~Gao, and N.~Sang, ``Jointly detecting and multiple people
  tracking by semantic and scene information,'' \emph{Neurocomputing}, vol.
  412, pp. 244--251, 2020.

\bibitem{lstm3}
S.~Wang, Y.~Zhou, J.~Yan, and Z.~Deng, ``Fully motion-aware network for video
  object detection,'' in \emph{Proceedings of the European conference on
  computer vision (ECCV)}, 2018, pp. 542--557.

\bibitem{lstm}
S.~Hochreiter and J.~Schmidhuber, ``Long short-term memory,'' \emph{Neural
  computation}, vol.~9, no.~8, pp. 1735--1780, 1997.

\bibitem{tracker}
P.~Bergmann, T.~Meinhardt, and L.~Leal-Taixe, ``Tracking without bells and
  whistles,'' in \emph{Proceedings of the IEEE/CVF International Conference on
  Computer Vision}, 2019, pp. 941--951.

\bibitem{tracker2}
W.~Yang, B.~Liu, W.~Li, and N.~Yu, ``Tracking assisted faster video object
  detection,'' in \emph{2019 IEEE International Conference on Multimedia and
  Expo (ICME)}.\hskip 1em plus 0.5em minus 0.4em\relax IEEE, 2019, pp.
  1750--1755.

\bibitem{predicting}
P.~Luc, N.~Neverova, C.~Couprie, J.~Verbeek, and Y.~LeCun, ``Predicting deeper
  into the future of semantic segmentation,'' in \emph{Proceedings of the IEEE
  International Conference on Computer Vision}, 2017, pp. 648--657.

\bibitem{bayesian}
A.~Bhattacharyya, M.~Fritz, and B.~Schiele, ``Bayesian prediction of future
  street scenes using synthetic likelihoods,'' in \emph{In International
  Conference on Learning Representations}, 2018.

\bibitem{segmenting}
H.-k. Chiu, E.~Adeli, and J.~C. Niebles, ``Segmenting the future,'' \emph{IEEE
  Robotics and Automation Letters}, vol.~5, no.~3, pp. 4202--4209, 2020.

\bibitem{vsaric2019single}
J.~{\v{S}}ari{\'c}, M.~Or{\v{s}}i{\'c}, T.~Antunovi{\'c}, S.~Vra{\v{z}}i{\'c},
  and S.~{\v{S}}egvi{\'c}, ``Single level feature-to-feature forecasting with
  deformable convolutions,'' in \emph{German Conference on Pattern
  Recognition}.\hskip 1em plus 0.5em minus 0.4em\relax Springer, 2019, pp.
  189--202.

\bibitem{warp}
J.~Saric, M.~Orsic, T.~Antunovic, S.~Vrazic, and S.~Segvic, ``Warp to the
  future: Joint forecasting of features and feature motion,'' in
  \emph{Proceedings of the IEEE/CVF Conference on Computer Vision and Pattern
  Recognition}, 2020, pp. 10\,648--10\,657.

\bibitem{predictive}
Z.~Lin, J.~Sun, J.-F. Hu, Q.~Yu, J.-H. Lai, and W.-S. Zheng, ``Predictive
  feature learning for future segmentation prediction,'' in \emph{Proceedings
  of the IEEE/CVF International Conference on Computer Vision}, 2021, pp.
  7365--7374.

\bibitem{luc2018predicting}
P.~Luc, C.~Couprie, Y.~Lecun, and J.~Verbeek, ``Predicting future instance
  segmentation by forecasting convolutional features,'' in \emph{Proceedings of
  the european conference on computer vision (ECCV)}, 2018, pp. 584--599.

\bibitem{sun2019predicting}
J.~Sun, J.~Xie, J.-F. Hu, Z.~Lin, J.~Lai, W.~Zeng, and W.-S. Zheng,
  ``Predicting future instance segmentation with contextual pyramid
  convlstms,'' in \emph{Proceedings of the 27th acm international conference on
  multimedia}, 2019, pp. 2043--2051.

\bibitem{apanet}
J.-F. Hu, J.~Sun, Z.~Lin, J.-H. Lai, W.~Zeng, and W.-S. Zheng, ``Apanet:
  Auto-path aggregation for future instance segmentation prediction,''
  \emph{IEEE Transactions on Pattern Analysis and Machine Intelligence}, 2021.

\bibitem{head1}
G.~Song, Y.~Liu, and X.~Wang, ``Revisiting the sibling head in object
  detector,'' in \emph{Proceedings of the IEEE/CVF Conference on Computer
  Vision and Pattern Recognition}, 2020, pp. 11\,563--11\,572.

\bibitem{head2}
Y.~Wu, Y.~Chen, L.~Yuan, Z.~Liu, L.~Wang, H.~Li, and Y.~Fu, ``Rethinking
  classification and localization for object detection,'' in \emph{Proceedings
  of the IEEE/CVF conference on computer vision and pattern recognition}, 2020,
  pp. 10\,186--10\,195.

\bibitem{ota}
Z.~Ge, S.~Liu, Z.~Li, O.~Yoshie, and J.~Sun, ``Ota: Optimal transport
  assignment for object detection,'' in \emph{Proceedings of the IEEE/CVF
  Conference on Computer Vision and Pattern Recognition}, 2021, pp. 303--312.

\bibitem{yu2020bdd100k}
F.~Yu, H.~Chen, X.~Wang, W.~Xian, Y.~Chen, F.~Liu, V.~Madhavan, and T.~Darrell,
  ``Bdd100k: A diverse driving dataset for heterogeneous multitask learning,''
  in \emph{CVPR}, 2020.

\bibitem{cordts2016cityscapes}
M.~Cordts, M.~Omran, S.~Ramos, T.~Rehfeld, M.~Enzweiler, R.~Benenson,
  U.~Franke, S.~Roth, and B.~Schiele, ``The cityscapes dataset for semantic
  urban scene understanding,'' in \emph{CVPR}, 2016.

\bibitem{caesar2020nuscenes}
H.~Caesar, V.~Bankiti, A.~H. Lang, S.~Vora, V.~E. Liong, Q.~Xu, A.~Krishnan,
  Y.~Pan, G.~Baldan, and O.~Beijbom, ``nuscenes: A multimodal dataset for
  autonomous driving,'' in \emph{CVPR}, 2020.

\bibitem{swish}
P.~Ramachandran, B.~Zoph, and Q.~V. Le, ``Swish: a self-gated activation
  function,'' \emph{arXiv preprint arXiv:1710.05941}, vol.~7, p.~1, 2017.

\bibitem{nonlocal}
X.~Wang, R.~Girshick, A.~Gupta, and K.~He, ``Non-local neural networks,'' in
  \emph{Proceedings of the IEEE conference on computer vision and pattern
  recognition}, 2018, pp. 7794--7803.

\bibitem{stn}
M.~Jaderberg, K.~Simonyan, A.~Zisserman \emph{et~al.}, ``Spatial transformer
  networks,'' \emph{Advances in neural information processing systems},
  vol.~28, pp. 2017--2025, 2015.

\bibitem{senet}
J.~Hu, L.~Shen, and G.~Sun, ``Squeeze-and-excitation networks,'' in
  \emph{Proceedings of the IEEE conference on computer vision and pattern
  recognition}, 2018, pp. 7132--7141.

\bibitem{flownet}
A.~Dosovitskiy, P.~Fischer, E.~Ilg, P.~Hausser, C.~Hazirbas, V.~Golkov, P.~Van
  Der~Smagt, D.~Cremers, and T.~Brox, ``Flownet: Learning optical flow with
  convolutional networks,'' in \emph{Proceedings of the IEEE international
  conference on computer vision}, 2015, pp. 2758--2766.

\bibitem{ggiou}
S.~Wu, J.~Yang, H.~Yu, L.~Gou, and X.~Li, ``Gaussian guided iou: A better
  metric for balanced learning on object detection,'' \emph{arXiv preprint
  arXiv:2103.13613}, 2021.

\bibitem{iou_balanced}
S.~Wu, J.~Yang, X.~Wang, and X.~Li, ``Iou-balanced loss functions for
  single-stage object detection,'' \emph{arXiv preprint arXiv:1908.05641},
  2019.

\bibitem{coco}
T.-Y. Lin, M.~Maire, S.~Belongie, J.~Hays, P.~Perona, D.~Ramanan,
  P.~Doll{\'a}r, and C.~L. Zitnick, ``Microsoft coco: Common objects in
  context,'' in \emph{European conference on computer vision}.\hskip 1em plus
  0.5em minus 0.4em\relax Springer, 2014, pp. 740--755.

\bibitem{linear}
P.~Goyal, P.~Doll{\'a}r, R.~Girshick, P.~Noordhuis, L.~Wesolowski, A.~Kyrola,
  A.~Tulloch, Y.~Jia, and K.~He, ``Accurate, large minibatch sgd: Training
  imagenet in 1 hour,'' \emph{arXiv preprint arXiv:1706.02677}, 2017.

\bibitem{adascale}
T.-W. Chin, R.~Ding, and D.~Marculescu, ``Adascale: Towards real-time video
  object detection using adaptive scaling,'' in \emph{In Proceedings of Machine
  Learning and Systems}, 2019.

\end{thebibliography}
\end{document}